\documentclass{article}

\usepackage{booktabs}
\usepackage{multirow}
\usepackage{subcaption}
\usepackage[table]{xcolor}
\usepackage{array}
\usepackage{hhline}
\usepackage{graphicx}
\usepackage{enumitem} 
\usepackage{wrapfig}
\usepackage{caption}
\definecolor{myblue}{HTML}{1167b1}
\definecolor{lightblue}{HTML}{BCD2E8}
\definecolor{myorange}{HTML}{ff9c59}
\definecolor{lightorange}{HTML}{ffdbbb}
\usepackage[preprint]{neurips_2025}

\usepackage[utf8]{inputenc} 
\usepackage[T1]{fontenc}    
\usepackage{url}            
\usepackage{booktabs}       
\usepackage{amsfonts}       
\usepackage{nicefrac}       
\usepackage{microtype}      

\usepackage[
    pagebackref,
    breaklinks,
    citecolor=uclablue,
    colorlinks=true,
]{hyperref}
\definecolor{uclablue}{rgb}{0.15, 0.45, 0.68}

\usepackage[most]{tcolorbox}

\title{\textsc{Embodied Web Agents}: Bridging Physical-Digital Realms for Integrated Agent Intelligence}

\author{
  Yining Hong* \quad
  Rui Sun* \quad
  Bingxuan Li† \quad
  Xingcheng Yao† \quad
  Maxine Wu† \quad
  Alexander Chien† \\
  \textbf{Da Yin} \quad
  \textbf{Ying Nian Wu} \quad
  \textbf{Zhecan James Wang} \quad
  \textbf{Kai-Wei Chang} \\ \\
  University of California, Los Angeles
}

\begin{document}

\maketitle

\begin{abstract}
AI agents today are mostly siloed — they either retrieve and reason over vast amount of digital information and knowledge obtained online; or interact with the physical world through embodied perception, planning and action — but rarely both. This separation limits their ability to solve tasks that require integrated physical and digital intelligence, such as cooking from online recipes, navigating with dynamic map data, or interpreting real-world landmarks using web knowledge. We introduce \textsc{Embodied Web Agents}, a novel paradigm for AI agents that fluidly bridge embodiment and web-scale reasoning. 
To operationalize this concept, we first develop the \textsc{Embodied Web Agents} task environments, a unified simulation platform that tightly integrates realistic 3D indoor and outdoor environments with functional web interfaces. Building upon this platform, we construct and release the \textsc{Embodied Web Agents} Benchmark, which encompasses a diverse suite of tasks including cooking, navigation, shopping, tourism, and geolocation — all requiring coordinated reasoning across physical and digital realms for systematic assessment of cross-domain intelligence.
Experimental results reveal significant performance gaps between state-of-the-art AI systems and human capabilities, establishing both challenges and opportunities at the intersection of embodied cognition and web-scale knowledge access. All datasets, codes and websites are publicly available at our project page \url{https://embodied-web-agent.github.io/}.
\end{abstract}

\section{Introduction}

\begin{figure}
    \centering
    \includegraphics[width=1\linewidth]{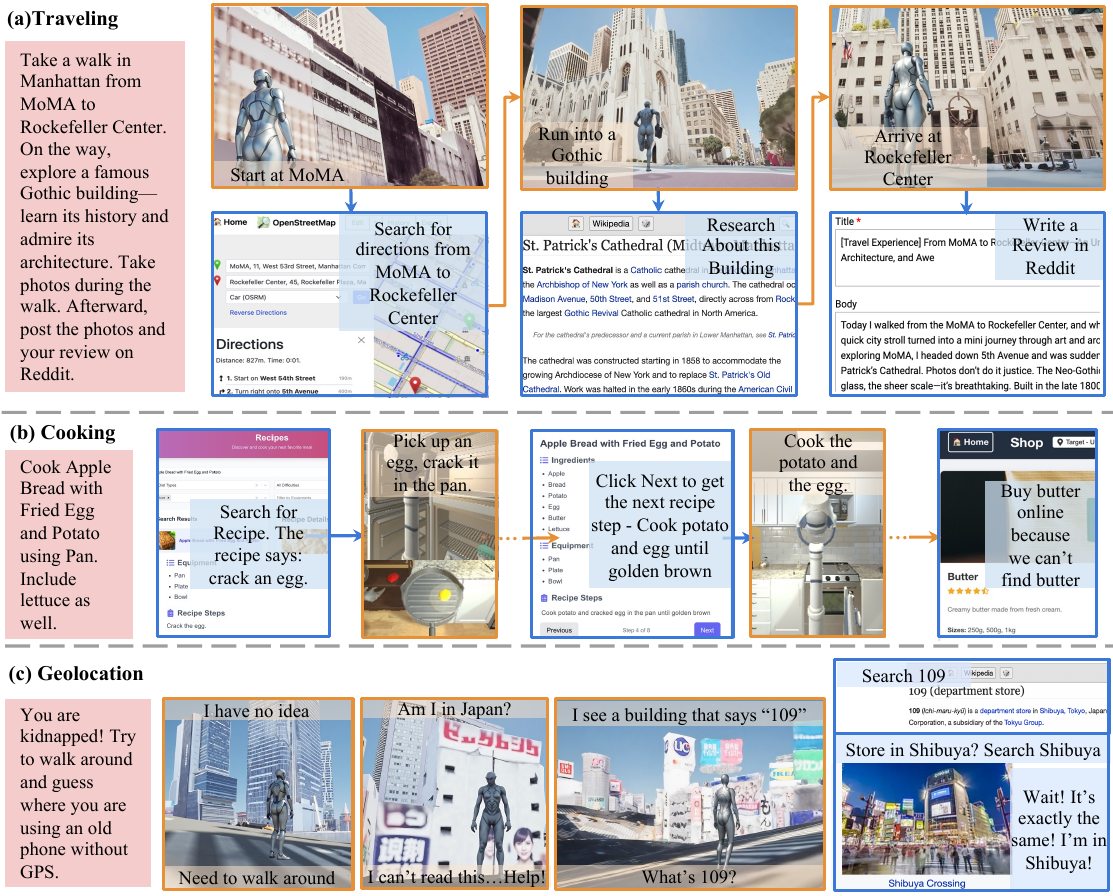}
    \caption{
        \textbf{Illustrative examples of our \textsc{Embodied Web Agents} conceptual paradigm, tasks and environments.} \textcolor{myblue}{Blue boxes and arrows} indicate web interaction / switching to the web respectively.  \textcolor{myorange}{Orange boxes and arrows} indicates acting in / switching to the embodied environment. We omit most intermediate actions due to the large number of interaction steps.}

    \label{fig:teaser}
\end{figure}
Recently, we have seen the proliferation of web agents capable of retrieving information online ~\citep{shi2017world, yao2022webshop, deng2023mind2web, webarena, koh-etal-2024-visualwebarena} — yet they remain confined to screens disembodied from the real world. Meanwhile, their physical counterparts — robots and embodied systems — navigate the world but with limited access to the Internet. What if the boundary between the digital and physical realms were shattered? What if web agents stepped out of the browser, with keys to perceive and act in the real 3D physical world, while physical robots autonomously tapped into the encyclopedic knowledge of the web?  As illustrated in Figure \ref{fig:teaser}, such agents would not only assess the ingredients in your kitchen, search for matching recipes online, shop for missing items, and cook your favorite dish for you; but also traverse historical landmarks, interpret architectural styles using both their own perception and Wikipedia, leave personalized reviews, and perhaps even return with a souvenir in hand. We, as humans, don't compartmentalize our intelligence into "physical-only" and "digital-only" modules — we fluidly move between realms. What if contemporary AI agents could likewise achieve the best of both worlds?

Building such agents \textit{goes far beyond a mere combination of isolated web and embodied systems}; it presents a set of deeply intertwined challenges.
The first is \emph{the perceptual grounding problem}: how can an agent link abstract digital instructions (e.g., "cook potato and egg until golden brown" as in Figure \ref{fig:teaser} (b)) with the high-dimensional data streams of the physical world (e.g., visually recognizing the transition of potatoes and eggs to a golden state through a series of embodied observations)? 
Addressing this requires embodied perception, where agents actively interpret their surroundings through movement, interaction, and multimodal sensing — continually acquiring feedback from their environment and aligning these observations with digital instructions.
The second challenge is \emph{cross-domain planning}: how should an agent decide when to shift between physical actions and digital information retrieval, particularly when information from one domain contradicts or supplements the other? For instance, the online map may suggest a path to visit Rockefeller Center, but real-world observation may reveal that the center is closed due to a protest, demanding a dynamic reevaluation of the agent’s plan. To navigate seamlessly between domains, agents must maintain a coherent and persistent representation that bridges physical and digital contexts — recalling physical experiences when operating online, and retrieving digital knowledge when acting in the world. Despite all these challenges, there remains a surprising lack of research targeting this level of integrated intelligence — both in terms of conceptual frameworks and benchmark development. As a result, progress in each domain often unfolds in isolation, with limited cross-pollination between the two paradigms.

To this end, we introduce \textsc{Embodied Web Agents} as a new conceptual paradigm of AI systems that unify physical embodiment with web-scale knowledge access — capable of perceiving and acting in the real world while reasoning over dynamic, unstructured information from the web. To operationalize this concept, we first develop the \textsc{Embodied Web Agents} task environments, a unified simulation platform that integrates realistic 3D environments with interactive web interfaces. This platform combines (1) indoor settings from AI2-THOR, (2) outdoor navigation in Google Earth, and (3)  web interfaces including Wikipedia, online stores, recipe websites, map services \textit{etc.}, enabling agents to interact seamlessly with both physical and digital spaces. Building upon this environment, we construct the \textsc{Embodied Web Agents} Benchmark, which encompasses approximately 1.5k tasks across multiple domains, including: (1) cooking tasks where agents match physical ingredients with online recipes; (2) navigation combining online maps with physical wayfinding; (3) shopping requiring coordination between in-store actions and online options; (4) tourism connecting physical landmarks with web information; and (5) geolocation determining position through embodied exploration and online research. Together, these tasks systematically test an agent's ability to bridge embodied perception, action, and web-based reasoning across varied contexts.

We conduct comprehensive experiments on our proposed \textsc{Embodied Web Agents} benchmark using several state-of-the-art LLM agent baselines, including GPT, Gemini, Qwen, and Intern models. Experimental results show that current LLM agents are far from satisfactory compared to human performances. 
A detailed breakdown and analysis of error types and their percentage contributions to task failures also reveal that current models predominantly struggle with cross-domain integration, not isolated capabilities.  For instance, these models encounter problems such as being trapped in a single environment and unable to switch to the other domain, or the misalignment of web instructions and embodied actions.
This further strengthens our position that embodied web agency presents unique challenges that cannot be studied through isolated physical or digital agents alone, as the key difficulties emerge precisely at the intersection where these domains are intertwined.

The key contributions of this paper can be summarized as follows. 
\begin{itemize}[noitemsep, topsep=0pt]
  \setlength\itemsep{0em}
  \setlength\parskip{0em}
  \setlength\parsep{0em}
\item We introduce \textsc{Embodied Web Agents} as a new conceptual paradigm for AI systems that integrate embodiment with web-scale information access — formalizing a class of agents capable of acting in the physical world while reasoning over unstructured digital content.

\item We develop the \textsc{Embodied Web Agents} task environments, a unified simulation platform that tightly integrates realistic 3D environments with interactive web interfaces, enabling agents to perform cross-domain tasks involving perception, action, and retrieval. 

\item We construct and release the \textsc{Embodied Web Agents} Benchmark, which encompasses a diverse suite of tasks across multiple domains including navigation, shopping, traveling, cooking and geolocation.

\item  We conduct in-depth empirical analysis of state-of-the-art LLM agents on our benchmark, revealing that our benchmark poses rigorous challenges for current LLM agents, and opens up a challenging new direction and testbed for future agents with integrated intelligence.
\end{itemize}

\section{Related Works}

\noindent \textbf{Web Agent Benchmarks}
Web agents are designed to navigate and interact with web environments to complete tasks following user instruction. Initial web agent evaluation benchmarks such as MiniWoB~\citep{shi2017world} and MiniWoB++~\citep{liu2018reinforcement} introduce a suite of diverse web navigation tasks on synthetic webpages. More recent benchmarks emphasize greater realism and task diversity. WebShop~\citep{yao2022webshop} simulates an e-commerce platform with numbers of products to evaluate agents' ability to search and make purchases, while Mind2Web~\citep{deng2023mind2web} provides a diverse collection of open-ended tasks across hundreds of real websites to assess general web navigation and interaction capabilities. Similarly, benchmarks like WebArena~\citep{webarena}, WebVoyager~\citep{he-etal-2024-webvoyager}, WebLINX~\citep{luweblinx}, and VisualWebArena~\citep{koh-etal-2024-visualwebarena} feature fully functional websites spanning multiple domains, enabling the evaluation of agents on long-horizon tasks in realistic, diverse environments. OVEN~\citep{hu2023opendomainvisualentityrecognition} challenges models to link images to specific Wikipedia entities given text queries. Beyond pursuing more realistic test environments, WorkArena~\citep{drouin2024workarena} requires agents to interact with enterprise software and perform tasks demanding higher expertise and comprehension. In this work, we explore a distinct yet important scenario where web browsing is integrated into the physical embodied world.

\noindent \textbf{Embodied Environments and Benchmarks}
Recent developments in environments and benchmarks have accelerated the research on embodied AI. Simulation platforms, such as AI2-THOR~\citep{ai2thor}, Habitat~\citep{habitat19iccv} and iGibson~\citep{shen2021igibson,li2022igibson}, enable agents to perform diverse interactive tasks in realistic indoor environments. Benchmarks like ALFRED~\citep{ALFRED20} and BEHAVIOR~\citep{srivastava2021behavior} provide a diverse suite of indoor tasks for embodied agents, requiring instruction understanding, long-horizon planning and manipulation in a closed environment. Additionally, Embodied Agent Inferface~\citep{li2024embodied} formalizes decision processes for LLM-based embodied agents and introduces fine-grained evaluation metrics for indoor embodied tasks. Efforts have also been made to extend the applicability of embodied agents to outdoor environments. A series of outdoor navigation benchmarks, such as StreetLearn~\citep{NEURIPS2018_streetlearn}, TouchDown~\citep{Chen_2019_touchdown,mehta-etal-2020-retouchdown}, RUN~\citep{paz-argaman-tsarfaty-2019-run}, have been introduced to evaluate the ability of embodied agents on vision-language navigation and spatial description resolution in urban street environments. \cite{Du2016SocialStreetView, Du2019Geollery} create immersive systems that integrate geo-tagged social media with 3D street-level environments to enhance virtual and augmented reality experiences for storytelling, tourism, and cultural exploration. \cite{yang2024virlgroundingvirtualintelligence} V-IRL is a platform for training and testing AI agents in realistic virtual environments to develop real-world skills. More outdoor related tasks such as geolocation prediction~\citep{haas2023pigeon} and map understanding~\citep{xing2025large} has also been proposed recently. In this work, we design a new benchmark encompassing a diverse set of embodied tasks within both indoor and outdoor environments. Different from previous works, our benchmark focuses on embodied tasks that require web access and interaction to be completed, a realistic scenario that is challenging and neglected in existing benchmarks. 

\noindent \textbf{Cross-Modal Agent Systems}
Cross-modal agent systems integrating vision, language and other modalities have been explored in both web and embodied environments. In web-based settings, ~\cite{he-etal-2024-webvoyager} builds a web agent powered by a large multimodal model that interacts with real-world websites following user instructions. ~\cite{lin2024showui} develops ShowUI, an efficient vision-laguage-action model for GUI agent. For embodied tasks, multimodal foundation models such as Gato~\citep{reed2022generalist}, PaLM-E~\citep{driess2023palme} and 3D-LLM~\citep{3dllm} have been developed to provide generalist policies in real world. In this work, we explore a new dimension for modal fusion in embodied agents, by integrating both embodied and web actions into one unified framework, to enable agents to perform more complex and diverse tasks with real-world applications.

\section{The \textsc{Embodied Web Agent} Task Environments}
\begin{figure}
    \centering
    \includegraphics[width=1\linewidth]{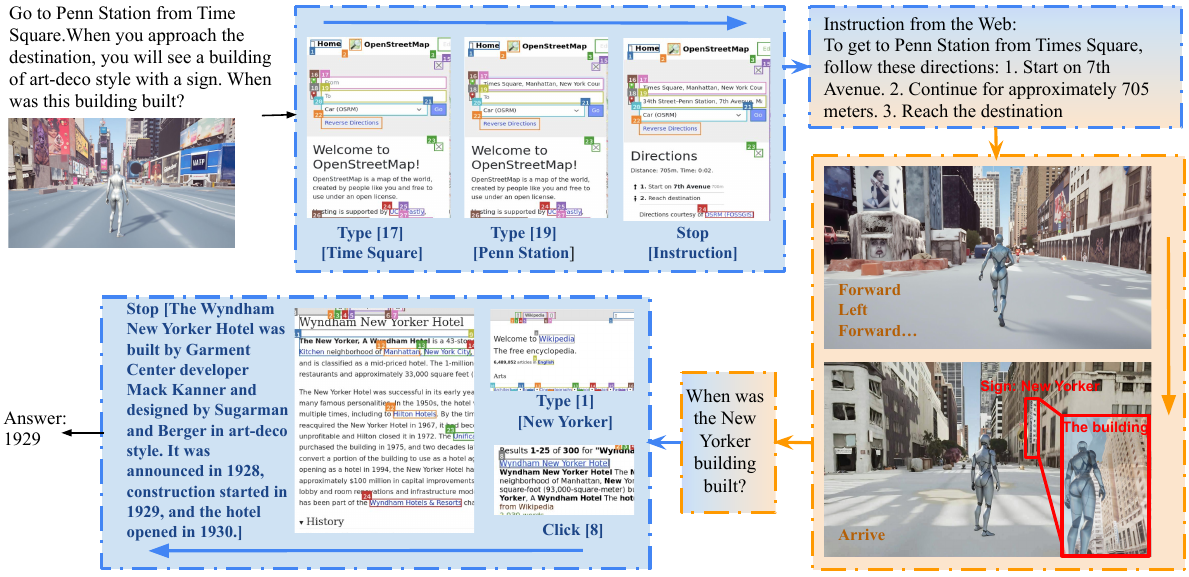}
    \caption{
        \textbf{An Exemplar Pipeline of completing a task in our \textsc{Embodied Web Agents} dataset.} \textcolor{myblue}{Blue boxes indicate web interaction}. \textcolor{myorange}{Orange boxes indicate embodied interaction}. Boxes with gradient colors indicate switching from one environment to the other. }

    \label{fig:web}
\end{figure}

Inspired by \cite{webarena}, our  environments are formalized as $E = \langle S, A, O, T \rangle$, where $S$ is the combined physical-digital state space, $A$ is the action space spanning both domains, and $O$ is the observation space comprising embodied input $o^e_t$ and web perception $o^w_t$. The deterministic transition function $T: S \times A \rightarrow S$ governs state evolution as agents select actions based on task specification, observations, and history. Task completion is measured by reward function $r(a_1^T, s_1^T)$ evaluating whether actions successfully fulfill intents like cooking dishes or reaching destinations.

Our task environments can be categorized into three parts: outdoor environment  (\ref{sec:outdoor}), indoor environment (\ref{sec:indoor}) and web environment (\ref{sec:web}). We show an example of interacting with and switching among the environments in Figure \ref{fig:web}, as well as the action spaces of all environments in Table \ref{tab:action}.
 
\begin{minipage}[b]{0.65\textwidth}
    \centering
    \scriptsize
    \begin{tabular}{|l|l|}
        \hline
        \textbf{Action} & \textbf{Explanation} \\
        \hline
        \rowcolor{lightorange}
        \multicolumn{2}{|c|}{\textbf{INDOOR ENVIRONMENT ACTIONS}} \\
        \hline
        \multicolumn{2}{|c|}{\textit{Agent Movement}} \\
        \hline
        \texttt{Teleport [obj]} & Teleport agent to a specific object \\
        \texttt{MoveAhead/Back/Left/Right} & Move agent in a cardinal direction \\
        \hline
        \multicolumn{2}{|c|}{\textit{Object Interaction}} \\
        \hline
        \texttt{PickupObject} / \texttt{PutObject [obj]} & Pick up or put held object \\
        \hline
        \multicolumn{2}{|c|}{\textit{Object State Changes}} \\
        \hline
        \texttt{OpenObject} / \texttt{CloseObject [obj]} & Open or close an object \\
        \texttt{SliceObject [obj]} & Slice an object \\
        \texttt{CookObject [obj]} & Cook an object \\
        \hline
        \multicolumn{2}{|c|}{\textit{Environment Switching}} \\
        \hline
        \texttt{switch\_environment [msg]} & Switch between web/embodied \\
        \hline
        \rowcolor{lightorange}
        \multicolumn{2}{|c|}{\textbf{OUTDOOR ENVIRONMENT ACTIONS}} \\
        \hline
        \texttt{Forward / Left / Right} & Move agent in outdoor environment \\
        \hline
        \rowcolor{lightblue}
        \multicolumn{2}{|c|}{\textbf{WEB ENVIRONMENT ACTIONS}} \\
        \hline
        \multicolumn{2}{|c|}{\textit{Page Operation Actions}} \\
        \hline
        \texttt{click [id]} & Click on an element with specific id \\
        \texttt{type [id] [content] [pr]} & Type content into field \\
        \texttt{scroll [direction]} & Scroll page up or down \\
        \texttt{hover [id]} / \texttt{press [key\_comb]} & Hover or simulate key press \\
        \hline
        \multicolumn{2}{|c|}{\textit{Tab Management \& URL Navigation Actions}} \\
        \hline
        \texttt{new\_tab} / \texttt{close\_tab} / 
 \texttt{tab\_focus} & Open, close or focus on a tab \\
        \hline
        \texttt{goto [url]} / \texttt{go\_back / forward} & Navigate to URL or go back/forward \\
        \hline
    \end{tabular}
    \captionof{table}{Action Spaces for All Environments}
    \label{tab:action}
\end{minipage}
\hfill
\begin{minipage}[b]{0.35\textwidth}
  \centering
  \includegraphics[width=0.9\linewidth]{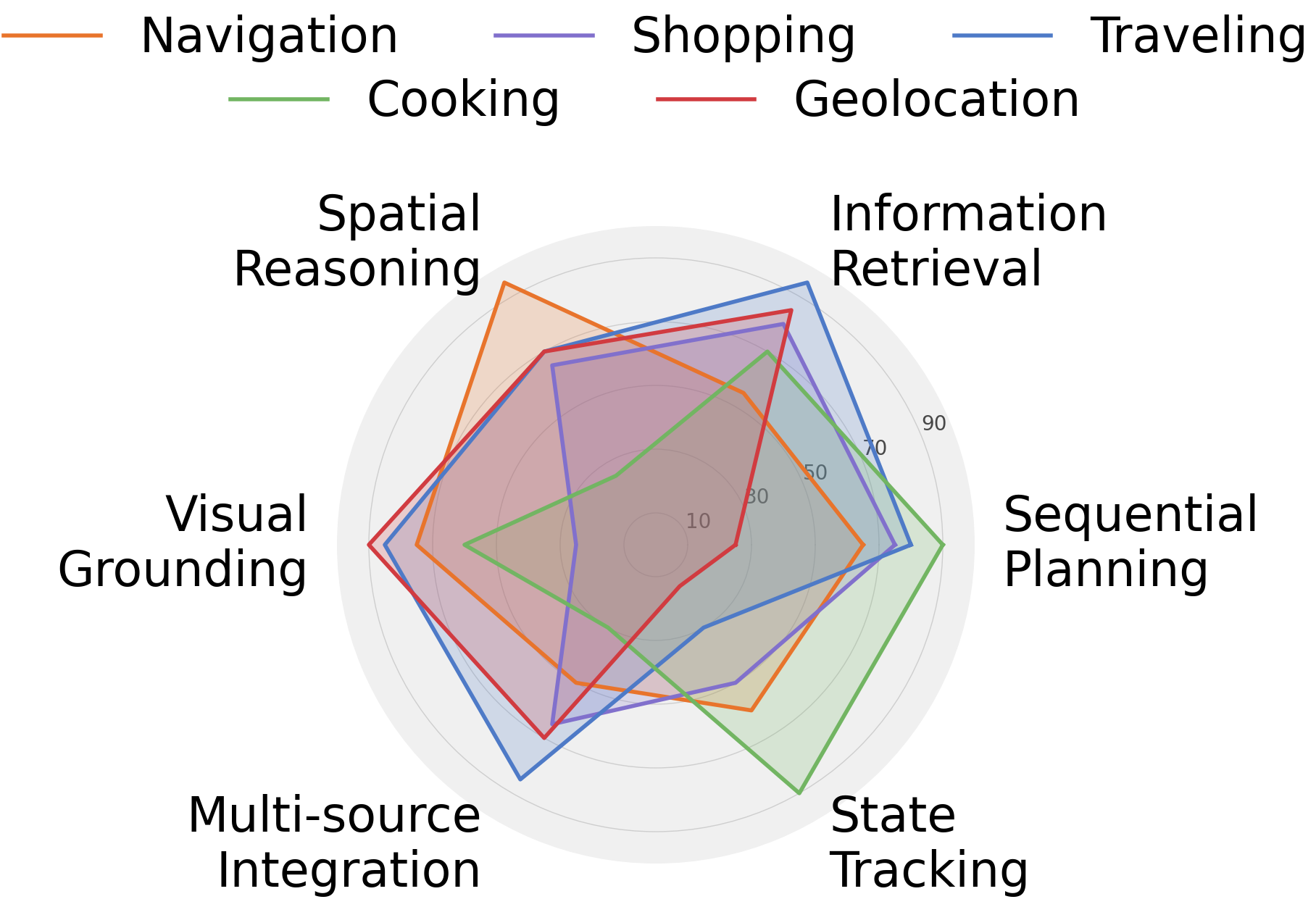} 
  \captionof{figure}{Importance of Different Capabilities Across Tasks}
  \label{fig:capability}
  \includegraphics[width=0.9\linewidth]{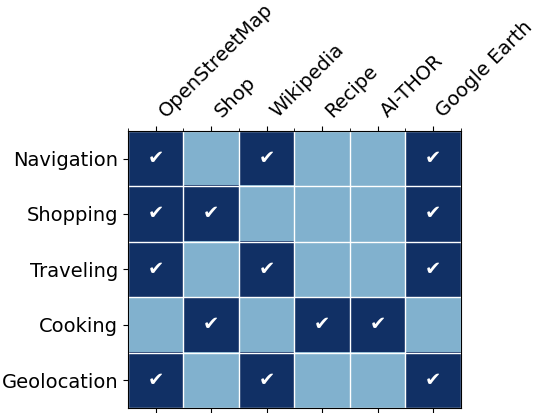}
  \captionof{figure}{Environments for Tasks}
  \label{fig:matrix}
\end{minipage}

\subsection{Outdoor Environment}\label{sec:outdoor}

The outdoor environment is constructed by leveraging the Google Street View and Google Earth API, which provides real-world, street-level observations captured by Google’s panoramic cameras. To build the outdoor environment, we select four cities (i.e., New York, Boston, Philadelphia, and Pittsburgh) with visually and structurally complex street layouts. Unlike synthetic or simulation-based environments, the visual data provided by Google is inherently more natural, noisy, and diverse, offering a more challenging and representative benchmark. Through API calls, we retrieve observations associated with specific geographic coordinates. These include panoramic images or standard-perspective images in cardinal directions. Alongside visual data, we also obtain: the GPS coordinates of each point, the heading / directional metadata between connected points, and the connectivity (adjacency) information across locations.  With these elements, we construct a navigation graph that underlies the outdoor environment. Formally, this environment can be described as an undirected graph $G = (V, E)$, where each node $v \in V$ represents a specific GPS coordinate, each edge $e \in E$ encodes a connection between two coordinates, including heading and distance, and each node is associated with four directional visual observations (north, east, south, west), represented as standard field-of-view images. Agents interact with the outdoor environment by observing these visual inputs, accessing the neighboring node set, and using heading information to reason about spatial transitions. Given navigation instructions (e.g., derived from web-based directions), the agent must determine which neighbor to move to at each step in order to reach a specified goal location, completing the navigation task through step-by-step decision making. This design closely mirrors real-world settings and introduces challenges that go beyond those posed by synthetic simulators. Compared to environments with simplified or rendered visuals, our outdoor environment demands stronger generalization and robustness from embodied agents, making it a more practical and realistic testbed for evaluating agent systems in open-world scenarios.

\subsection{Indoor Environment}\label{sec:indoor}

The indoor task environment utilizes AI2-THOR~\citep{ai2thor}, a photorealistic 3D indoor simulation platform. The environment provides highly accurate and interactive kitchen scenes containing fresh ingredients, cooking equipment, storage containers, and kitchen appliances. Agents can observe ingredient states, manipulate objects, and monitor cooking progress through visual perception. Objects are tracked with properties and states, including boolean flags (e.g., isSliced, isCooked), location information (e.g., parentReceptacles), and more, all of which dynamically update as agents execute physical actions like chopping or mixing, instructed by online recipes. A specialized state evaluator compares the current kitchen state against ideal target states, measuring task completion by checking whether objects have achieved desired states and spatial arrangements. 

\subsection{Web Environment} \label{sec:web}

The web environment consists of five functional websites, each supporting different aspects of agent interaction across both indoor and outdoor scenarios. The websites are implemented with a React.js frontend structured using modular components and state management, and a FastAPI backend that exposes asynchronous RESTful APIs for data serving and user interaction. The homepage serves as the central navigation hub, linking to all other task-specific websites and maintaining contextual continuity across interactions. The recipe website we built allows users to browse, search, and filter cooking recipes based on ingredients, dietary preferences, or cuisine types. The shopping website built from scratch enables management of a shopping cart, ingredient lookup, and simulated checkout processes. It facilitates task flows involving item selection, inventory reasoning, and purchasing. We also adapt several websites from the WebArena benchmark~\citep{webarena}. The OpenStreetMap site offers an interactive map for location search, address lookup, and exploration of geographic entities. The Wikipedia site presents richly interlinked encyclopedic content for information-seeking, entity linking, and multi-hop reasoning across documents. These websites are modified slightly to ensure smooth integration with the homepage. All websites are public and can be reached via \url{http://98.80.38.242:1220/}.  We also include more details and screenshots of the web environment in the Supplementary Material.


\section{The \textsc{Embodied Web Agents} Benchmark Construction}


In this section, we describe how we construct our \textsc{Embodied Web Agents} benchmark. We will cover 5 domains of tasks: Navigation, Shopping, Traveling, Cooking and Geolocation. We show examples of the tasks in Figure \ref{fig:teaser}, and a full pipeline of completing a task in Figure \ref{fig:web}. Figure \ref{fig:capability} summarizes the required level of each capability for successful task completion across domains, and Figure \ref{fig:matrix} shows which environments are utilized in different tasks.


\noindent \textbf{Navigation}
Building upon the Outdoor Environment described in $\S$~\ref{sec:outdoor}, our navigation tasks evaluate an agent's spatial reasoning ability to reach destinations based on web-sourced directions. We use the OpenStreetMap website in $\S$~\ref{sec:web} to ensure reproducibility and consistent web interaction. To create diverse navigation scenarios, we prompt GPT-4o-mini to generate geographic coordinates across the aforementioned cities. These coordinates serve as either the start or end points of a task, and the graph structure centered around each point can be developed using our outdoor environment. During the prompting process, we also generate initial task instructions tied to the obtained coordinates.
After identifying start or end points, we locate the corresponding counterparts using node adjacency relationships in the outdoor graph, forming a path within the environment. For evaluation purposes, we compute the shortest path using Dijkstra's algorithm as our ground-truth trajectory.

Navigation tasks require bidirectional interaction between web and embodied domains. The agent must input origin and destination into the map website to obtain directions, then ground these instructions in the embodied environment through turning actions and movements. Our benchmark includes 144 navigation tasks, each requiring both web interaction and embodied navigation. Since VLM-generated locations may have connectivity issues or misalignments with actual map data, we conduct human verification for all tasks to ensure their correctness and validity.

\noindent \textbf{Shopping}
In real life, when buying products, we typically compare prices online, decide where to purchase based on pricing and store location information, place an order online, and then visit a physical store for pickup. Our shopping tasks evaluate the agent’s ability to handle both online shopping and embodied environment interactions. The agent must place orders through our self-hosted shopping website dicussed in $\S$~\ref{sec:web} (\url{http://98.80.38.242:1207/}), obtain store locations, and navigate in the outdoor environment to the correct store for pickup using the directions by OpenStreetMap; alternatively, it may also first navigate to a store and then place the order online. 

In our benchmark, we simulate four stores located in distinct areas of Manhattan, New York. Our website lists a variety of items with product names, images, prices, and store information including distance and store name. The agent needs to \textit{weigh both the price of the item and the store's location to make an optimal decision}, ultimately grounding web information into the embodied environment and navigating to the store for the selected item. To generate diverse scenarios, we design multiple templates with different items and user intents, which are listed in detail in our Supplementary Material. We also test the agent's ability to retrieve information across multiple browser tabs—\textit{e.g.}, requiring the agent to complete a purchase, return to the homepage, switch to a map website, and search for directions before embodied navigation. Some complex tasks require multiple rounds of web interaction and physical navigation within a single shopping scenario, testing agents' multi-source integration and sequential planning abilities. In total, our dataset contains 216 shopping tasks.

\noindent \textbf{Traveling}
Inspired by how people consult web resources while traveling to navigate the physical world more effectively, we include traveling as a primary benchmark task. Using our custom-built outdoor environment and a pipeline similar to navigation tasks, we prompt a VLM to generate starting points, destinations, and initial task instructions, which we then refine into detailed, context-appropriate versions. Unlike pure navigation tasks that focus on following map directions and resolving map-reality inconsistencies, traveling tasks emphasize richer interaction between web resources and the embodied environment. For instance, when an agent encounters a significant landmark during navigation, as shown in Figure \ref{fig:teaser} (a) when it runs into a Gothic building, it may query Wikipedia to retrieve relevant information about that location. The agent is also expected to explore different architectural styles or historical landmarks, and ground Wikipedia descriptions to physical observations (\textit{e.g.,} grounding the text descriptions of appearances of a Gothic building to the actual observation of the building). Web interactions in traveling tasks extend beyond map reading to include diverse informative sites, creating scenarios with multiple intertwined interactions between digital and physical domains. Our benchmark includes 110 traveling tasks, each requiring fluid movement between embodied navigation and web-based information retrieval.



\noindent \textbf{Cooking}
As described in $\S$ \ref{sec:indoor}, we use AI2-THOR as our indoor environment. To generate embodied cooking tasks for execution, we begin by identifying all ingredients available in the AI2-THOR kitchen scenes. We then manually search online for recipes that include these ingredients. Since online recipes are often noisy and may not align with the constraints of the AI2-THOR environment, we use Claude to refine them. Claude is guided by a predefined set of allowable agent actions in AI2-THOR environment to ensure the resulting recipes are executable. To increase task difficulty, we introduce confounders for most of the recipes by including pairs of recipes with the same name but differing in difficulty level, dietary type, ingredients used, or required cooking equipment. The users can filter out recipes based on these constraints by filter bars below the search bar (as in \url{http://98.80.38.242:1206/}). The next step is to curate a set of tasks based on collected recipes. For each scene, we retrieve recipes that match the available ingredients. The task instruction asks the agent to cook the corresponding dish. When a confounder exists for a given recipe, we introduce additional constraints — \textit{e.g.}, “Diet type is vegetarian,” “Use a tomato,” — to disambiguate between recipe variants. If an ingredient does not exist in the scene, the agent is expected to go online to shop for it. The cooking tasks evaluate the agent's capability to perform long-trajectory planning in the indoor environment, and continuously check if the states match with the web instruction in the process. Our benchmark contains in total 911 cooking tasks. An exemplar task is in Figure \ref{fig:teaser} (b).

\noindent \textbf{Geolocation}
Geolocation is a classic computer vision task~\cite{hays2008im2gps}, where models predict geographic coordinates of given images. Instead of treating it purely as a conventional vision problem, we reinterpret it based on its inherent characteristics as an embodied geolocation task. Inspired by the design of \cite{geoguessr}, we move away from the single-image input setting and treat the model as an agent situated in an embodied environment. The agent is allowed to explore the outdoor environment we construct and ultimately output its estimated location. During exploration, the agent interprets storefront texts, visual cues, and street-view observations while accessing web information when needed to supplement its observations. The agent explores these environments freely, performing web interactions when additional information is needed. The task concludes when the agent has either 1) explored all possible positions or 2) collected sufficient information to confidently predict its location. This framework unifies embodied navigation, web-based reasoning, and visual grounding into a cohesive geolocation task. Our data collection is adapted from \cite{huang2025vlms}, focusing on examples from existing geolocation datasets where models typically fail. We select coordinates where we hypothesize web information may improve prediction accuracy, then construct environments centered on these points using Google API. Geolocation evaluates the visual grounding ability of agents. An example is shown in Figure \ref{fig:teaser} (c). We collect 142 such data.

\section{Experiments}

In this section, we first introduce baseline LLM agents ($\S$ \ref{sec:agent}) and evaluation metrics ($\S$ \ref{sec:metrics}) we use for  experiments. We then perform result analysis ($\S$ \ref{sec:result}) on our \textsc{Embodied Web Agents} benchmark. We group the results of Navigation, Shopping and Traveling together as they are all related to outdoor planning. Please refer to the Supplementary Material for more experimental results, experimental setup, LLM prompts, qualitative examples and error cases, as well as more analyses.

\subsection{Baseline LLM Agents} \label{sec:agent}

We evaluate four LLMs as our baseline agents: GPT-4o, Gemini 2.0 Flash, Qwen-VL-Plus, and InternVL2.5-latest. GPT-4o is OpenAI’s state-of-the-art multimodal model with strong performance in visual reasoning and real-time interaction. Gemini 2.0 Flash, by Google DeepMind, is optimized for speed and efficiency while maintaining robust vision-language capabilities. Qwen-VL-Plus, from Alibaba’s DAMO Academy, offers fine-grained image-text understanding. InternVL2.5-latest, developed by Shanghai AI Lab, excels in spatial and semantic reasoning.

\subsection{Evaluation Metrics}\label{sec:metrics}

To comprehensively assess agent performance across physical and digital domains, we employ four evaluation metrics for outdoor planning and cooking: \textbf{Overall Accuracy} measures the success of complete cross-domain task execution, requiring both successful web task completion (reaching the terminal web state) and fulfillment in the embodied environment, representing holistic task completion that necessitates seamless integration of both domains; \textbf{Web-only Accuracy} evaluates the ability to successfully complete the web portion of a task, such as reaching the final step of a recipe, isolating digital domain independent of physical execution; \textbf{Embodied-only Accuracy} assesses an agent's ability to achieve all required physical state conditions in the embodied environment, such as properly slicing  ingredients, or navigating to a desired place, measuring physical domain proficiency; and \textbf{Overall Completion Rate} represents the proportion of task progress achieved, indicating how much of the required state conditions have been fulfilled relative to the total task objectives.

\subsection{Result Analysis}\label{sec:result}

\begin{table}[h]
\centering
\small
\begin{tabular}{c|c|c|*{5}{c}}
\toprule
\multicolumn{3}{c|}{\textbf{Task / Metric}} & \textbf{GPT} & \textbf{Gemini} & \textbf{Qwen} & \textbf{Intern} & \textbf{Human} \\
\midrule
\midrule
\multirow{16}{*}{\rotatebox{90}{\textbf{Outdoor Tasks}}} 
& \multirow{4}{*}{\textbf{Navigation}} 
& Overall Accuracy & 34.72 & 30.56 & 15.97 & 13.19 & 90.28 \\
& & Overall Completion Rate & 52.08 & 48.96 & 36.81 & 26.04 & 91.32 \\
& & Web-only Accuracy & 69.44 & 67.36 & 57.64 & 38.89 & 92.36 \\
& & Embodied-only Accuracy & 48.61 & 46.53 & 31.25 & 23.61 & 90.97 \\
\hhline{~|~|------}
& \multirow{4}{*}{\textbf{Shopping}} 
& Overall Accuracy & 25.46 & 23.61 & 13.89 & 10.65 & 92.59 \\
& & Overall Completion Rate & 31.94 & 30.56 & 18.52 & 14.35 & 93.52 \\
& & Web-only Accuracy & 39.35 & 37.50 & 23.15 & 17.13 & 93.06 \\
& & Embodied-only Accuracy & 34.26 & 32.41 & 17.59 & 12.96 & 93.98 \\
\hhline{~|~|------}
& \multirow{4}{*}{\textbf{Traveling}} 
& Overall Accuracy & 30.91 & 25.45 & 11.82 & 9.09 & 91.82 \\
& & Overall Completion Rate & 50.91 & 48.18 & 34.55 & 20.91 & 93.64 \\
& & Web-only Accuracy & 57.27 & 53.64 & 41.82 & 25.45 & 94.55 \\
& & Embodied-only Accuracy & 47.27 & 44.55 & 29.09 & 19.09 & 92.73 \\
\bottomrule
\end{tabular}
\vspace{1mm}
\caption{\textbf{Model Performance Across Different Outdoor Tasks.} There is a huge performance gap between LLM agents' performances and human performances.}
\label{tab:outdoor_performance}

\end{table}

\begin{table}[h]
\centering
\small

\begin{tabular}{c|*{4}{c}|*{4}{c}|c}
\hline
\multirow{2}{*}{\textbf{Metric}} & \multicolumn{4}{c|}{\textbf{Vision}} & \multicolumn{4}{c|}{\textbf{Text}} & \multirow{2}{*}{\textbf{Human}} \\
\cline{2-9} &
\textbf{GPT} & \textbf{Gemini} & \textbf{Qwen} & \textbf{Intern} & \textbf{GPT} & \textbf{Gemini} & \textbf{Qwen} & \textbf{Intern} & \\
\hline
Overall Acc & 5.4 & 4.1 & 0.6& 0.0& 6.4 & 5.8 & 1.5 & 0.4 & 77.08\\
Completion Rate & 40.26& 35.62& 15.91& 9.73& 39.16 & 38.92 & 17.20 & 10.02 & 85.37\\
Web Acc & 59.71& 47.74& 28.65& 10.64& 57.08 & 62.23 & 35.89 & 15.58 & 100 \\
Embodied Acc & 8.7& 6.1& 2.2& 0.9& 10.5 & 8.2 & 4.1 & 1.3 & 77.08\\
\hline
\end{tabular}

\vspace{1mm}
\caption{\textbf{Model Performance for Cooking Task.} The models achieve inferior overall accuracies.}
\label{tab:indoor_performance}

\end{table}

\noindent \textbf{Outdoor Planning}
For outdoor planning, we use GPT-4o-mini alongside Gemini 2.0 Flash, Qwen-VL-Plus, and InternVL2.5-latest to evaluate performance across navigation, shopping, and traveling tasks (Table~\ref{tab:outdoor_performance}). For web observation, we follow the setting of VisualWebArena. We observe that: 1) GPT-4o-mini consistently leads across all metrics, with the highest accuracy in navigation (34.72\%), shopping (25.46\%), and traveling (30.91\%), though still well below human performance. Gemini follows closely behind, while Qwen and Intern lag behind. 2) Web-only accuracy exceeds embodied-only accuracy for all outdoor tasks, suggesting models handle digital information more effectively than physical navigation. 3) Generally, completion rates are satisfactory, while overall accuracies are very low across all tasks. This indicates models can execute parts of complex tasks but struggle with consistent cross-domain reasoning over longer sequences. 4) From task perspective, shopping and traveling involve richer interactions between the embodied environment and the web than navigation, and each task spans longer steps. As a result, the overall accuracy for shopping and traveling is noticeably lower than for navigation. This highlights the difficulty of cross-environment tasks, particularly those that are lengthy and involve multiple steps, for current models.

\noindent \textbf{Cooking}
For cooking, we implement two distinct approaches: vision-based and text-based. Our vision-based implementation draws inspiration from VisualWebArena, utilizing screenshot images of websites enhanced with Set-of-Marks (SoM) annotations that highlight interactive elements. For embodied observations, we provide first-person visual perspectives from the agent's viewpoint within the AI2-THOR environment. The text-based implementation follows WebArena's methodology, representing web content through accessibility trees that capture the semantic structure of websites in textual form. For embodied observations, we extract structured scene graphs directly from AI2-THOR, providing explicit object relationships and states. We use Qwen-PLUS and InternLM-latest for Qwen and Intern models without vision. 

Table~\ref{tab:indoor_performance} presents performance metrics for various models on the cooking task, comparing vision-based and text-based approaches against human performance. A substantial performance gap exists between AI models and humans, with the best model (text-based GPT-4o) achieving only 6.4\% overall accuracy compared to humans' 77.08\%. Text-based models using structured scene graphs consistently outperform their vision-based counterparts using first-person views, suggesting current models struggle to ground visual observations effectively in cooking contexts.
GPT-4o and Gemini-2.0-Flash demonstrate substantially stronger performance than Qwen-VL-Plus/Qwen-PLUS and InternVL/InternLM across both modalities. Notably, similar to outdoor performances, all models perform significantly better on web-only tasks compared to embodied-only tasks, revealing that while current models can navigate recipe websites effectively, they struggle with physical execution requiring object manipulation and state tracking.
Despite low overall accuracy, models achieve moderate completion rates, indicating partial task success but failure in full cross-domain integration. 

\noindent \textbf{Geolocation} For geolocation tasks, we benchmark against FairLocator~\citep{huang2025vlms}, a study analyzing VLM performance on GeoGuessr using Google Street View images. As shown in Table \ref{tab:geoguessr}, the embodied web agent, capable of active exploration and web information access, significantly outperforms the passive baseline, particularly in identifying finer-grained locations like cities and streets. We observe consistent improvements across all models when moving from the baseline to embodied setting, suggesting the performance gains are model-agnostic. Interestingly, we also find that even when the retrieved Wikipedia search results are noisy or uninformative, the act of querying itself often helps the agent reason more confidently. This indicates that formulating search queries may serve as a form of self-supervision.
This substantial improvement underscores the potential of integrating embodied and web domains to enhance performance across numerous real-world tasks, warranting further investigation.


\begin{table}[h]
\centering
\small
\begin{tabular}{c|c|c|c|c|c|c}
\toprule
\multicolumn{2}{c|}{\textbf{Setting / Model}} & \textbf{Continent} & \textbf{Country} & \textbf{City} & \textbf{Street} & \textbf{All} \\
\midrule
\midrule
\multirow{6}{*}{\rotatebox{90}{\textbf{Geolocation}}} 
& \textbf{FairLocator}   &      &       &       &       &       \\
& \quad GPT-4o-mini                        & 90.85 & 81.69 & 73.24 & 1.41 & 1.41 \\
& \quad Gemini-2.0-Flash             & 93.66 & 85.92 & 78.17 & 0.70 & 0.70 \\
& \quad Qwen-VL-Plus                 & 76.06 & 58.45 & 45.07 & 0.70 & 0.00 \\
& \quad InternVL2.5-Latest                 & 77.46 & 62.68 & 52.11 & 1.41 & 1.41 \\
\hhline{~|------}
& \textbf{Embodied Web Agent}         &      &       &       &       &       \\
& \quad GPT-4o-mini                       & 97.18 & 90.85 & 85.21 & 3.52 & 3.52 \\ 
& \quad Gemini-2.0-Flash             & 97.18 & 94.37 & 85.21 & 4.23 & 4.23 \\
& \quad Qwen-VL-Plus                 & 80.28     & 69.01      & 49.30      & 0.00     &  0.00     \\
& \quad InternVL2.5-Latest                 & 93.62 & 77.30 & 57.45 & 2.13 & 1.42 \\

\bottomrule
\end{tabular}

\vspace{1mm}
\caption{\textbf{Model performance for geolocation task.} All models performed much better when predicting after interactively exploring the environment and querying the web than just using static images.}
\label{tab:geoguessr}

\end{table}

\subsection{Error Analysis}

\begin{wrapfigure}{R}{0.55\textwidth}
  \centering
  
  \includegraphics[width=0.55\textwidth]{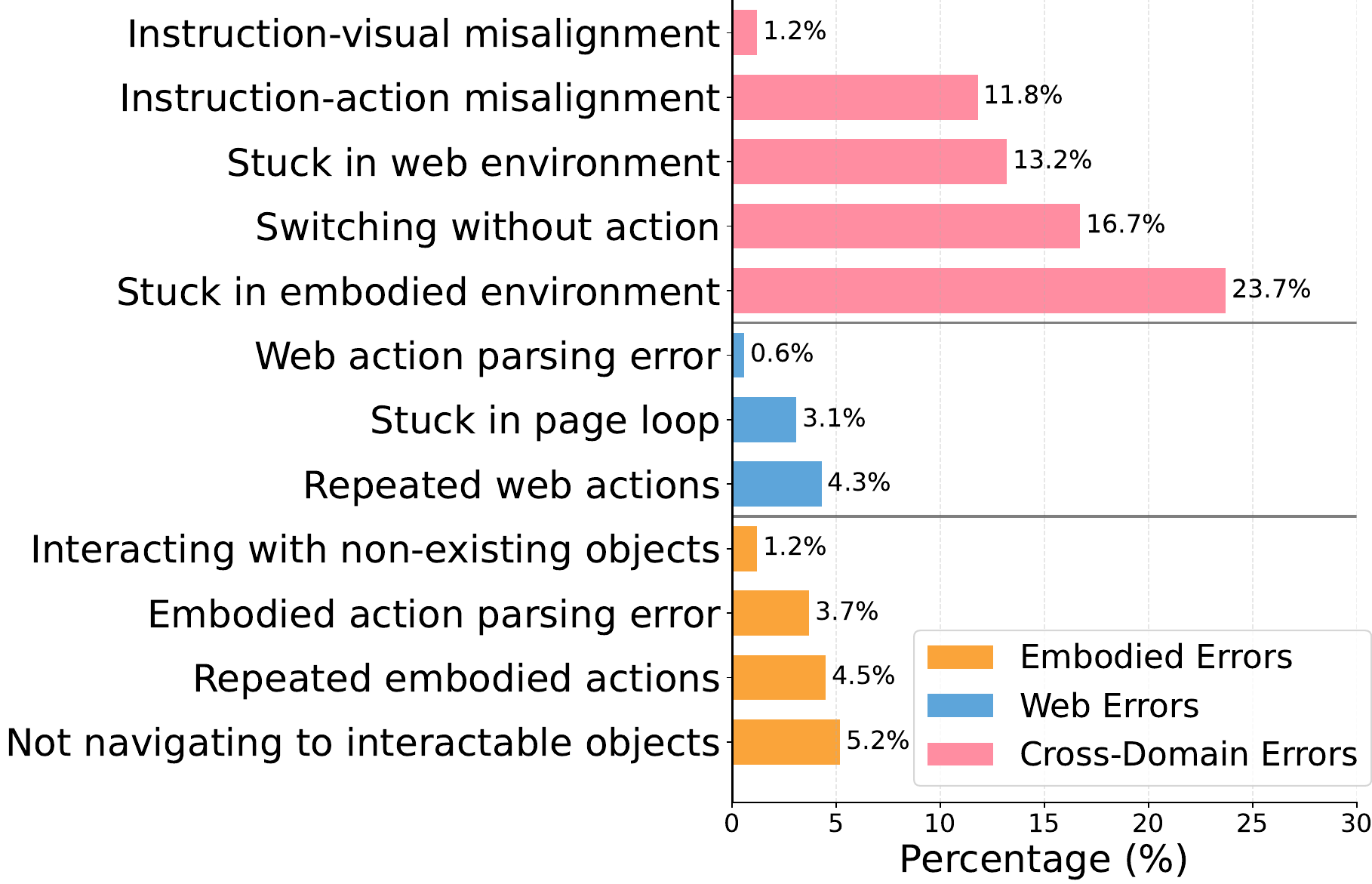}
  \caption{\textbf{Error Analysis for Cooking Tasks.} We can see that the majority of errors are cross-domain errors. }
  \label{fig:error_analysis}
\end{wrapfigure}

Figure \ref{fig:error_analysis} presents a detailed breakdown of error types and their percentages that contribute to task failures in cooking tasks when using GPT-4o. Our analysis reveals that the primary challenges in embodied web agents lie not in isolated capabilities, but in their integration. While embodied errors (14.6\%) and web errors (8.0\%) occur, cross-domain errors (66.6\%) overwhelmingly dominate the failure landscape — confirming that the critical bottleneck emerges at the intersection where physical and digital domains meet. The most prevalent failure pattern involves agents becoming trapped in single-domain cycles. In 23.6\% of failures, agents get stuck in the embodied environment, repeatedly executing irrelevant physical actions without returning to the web for the next step. Similarly, in 13.2\% of cases, agents remain fixed in web environments, endlessly clicking "next" through recipe pages without initiating cooking actions. In addition, agents often switch between environments without meaningful action (16.7\%) or suffer from instruction-action misalignments (11.8\%), such as slicing lettuce when a recipe instructs "slice the apple". Web interaction failures manifest as agents getting stuck in page loops (3.1\%) or performing identical actions repeatedly (4.3\%). In the embodied domain, agents fail to navigate to interactable objects (5.2\%) or execute repeated actions (4.5\%). These isolated domain errors are far less frequent than cross-domain integration failures, explaining why LLM agents achieve only 6.4\% overall accuracy despite moderate performance on single-domain tasks. This confirms that embodied web agency presents unique challenges requiring focused research on mechanisms that bridge physical and digital reasoning.

\section{Conclusion}

In this paper, we introduced \textsc{Embodied Web Agents}, a new paradigm for AI research that bridges the artificial divide between physical and digital intelligence. Through our comprehensive benchmark spanning cooking, navigation, shopping, tourism, and geolocation tasks, we demonstrate that current AI systems face significant challenges in fluidly integrating embodied perception with web-based information retrieval. These findings establish a foundation for future research in integrated intelligence systems, highlighting the need for developing AI agents that can seamlessly traverse physical and digital worlds. A limitation is our reliance on simulated agents, which may not fully capture the complexity and unpredictability of physical-digital interactions of real robots.


\bibliographystyle{plainnat}
\bibliography{references}

\clearpage
\newpage

\renewcommand{\thesection}{\Alph{section}}  

\addtocontents{toc}{\protect\setcounter{tocdepth}{3}} 
\addtocontents{toc}{\protect\setcounter{section}{0}}  
\appendix

{
    \hypersetup{linkcolor=black}
    \tableofcontents
}
\clearpage

\section{Contribution Statement}
Yining Hong is responsible for coming up with the idea; overall organization of the team (meetings; pointing out research directions; divison of responsibilities; reaching out to potential collaborators etc.); all the data, codes and experiments of indoor cooking; the majority of paper writing; creating demo videos.

Rui Sun implemented interactions between the outdoor environment and the web, made corrections to the environment, collected data for outdoor tasks (i.e., navigation, traveling, and shopping), and validated the data collection process. Rui also wrote part of the paper about outdoor tasks. He also wrote detailed instructions for implementing the Geolocation tasks for Maxine and Alexander. 

Bingxuan Li took care of designing and implementing all the web environments used in this paper. Bingxuan also made the showcase website of this paper, and wrote the web development part of the paper.

Xingcheng Yao built the basic outdoor environment using the Google Street View API, which lays a good foundation for further development. Xingcheng also wrote part 2 and part 3 of related works.

Maxine Wu and Alexander Chien were responsible for the entire Geolocation section. Maxine implemented the baseline pipeline, contributed to the design of evaluation metrics, performed error analysis, and created the demo videos. Alexander implemented and contributed to the design of the embodied environment and the web interaction system for the final pipeline, curated the dataset, and wrote the corresponding sections of the paper. Maxine also adapted the pipeline to support different models. Both authors performed experiments in both the baseline and the embodied settings.

Da Yin came up with first-step instructions of the Geolocation task. He also wrote the web agent part of the related works.

Yingnian Wu, Zhecan James Wang and Kai-Wei Chang took the advising roles. Specifically, Prof. Wu provided initial insights on agent planning. 
Zhecan James Wang helped sort out the meeting notes and discussion results into documents; provided ideas on task design; helped coordination among people. 
Prof. Chang scheduled biweekly meetings with the team, gave valuable advice and pointed out valuable research directions, as well as helped polish the paper.  

\section{Broader Impacts}
Our \textsc{Embodied Web Agents} research presents both opportunities and challenges for society. On the positive side, agents that bridge physical and digital domains could enhance accessibility for individuals with mobility limitations, support contextualized learning environments, and improve emergency response through integrated information access. However, several risks warrant attention. First, these agents may exhibit "dual-domain hallucination," where errors propagate across physical and digital realms, compounding misinformation. Second, systems that connect physical environments with web platforms introduce novel privacy concerns beyond those in either domain alone. 

To mitigate these concerns, our benchmark provides transparent evaluation protocols that can identify cross-domain errors. We designed our environments as simulations that don't interact with real-world systems, limiting immediate risks while providing valuable research insights. By releasing our benchmark to the research community, we aim to encourage the development of more robust embodied web agents with improved error detection mechanisms before deployment in real-world settings.

\section{Dataset Statistics}
\begin{figure}[htbp]
    \centering
    \includegraphics[width=1\linewidth]{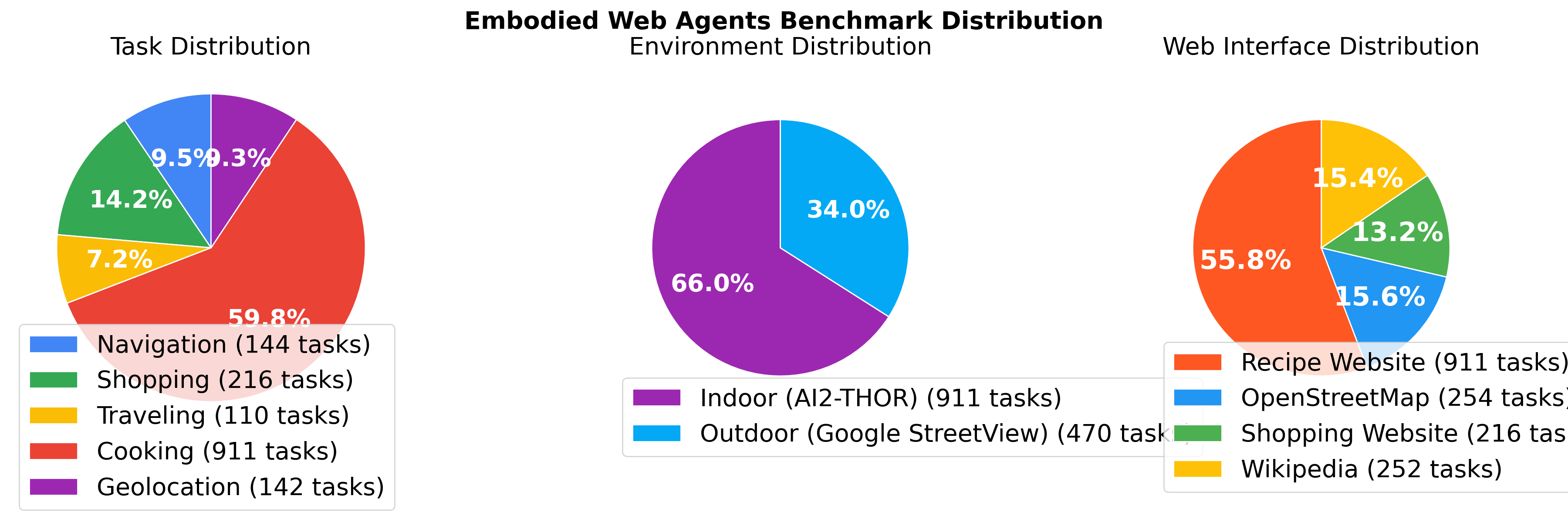}
    \caption{Task, scene and web distributions of our data}
    \label{fig:dataset_details}
\end{figure}

\begin{figure}[htbp]
    \centering
    \includegraphics[width=1\linewidth]{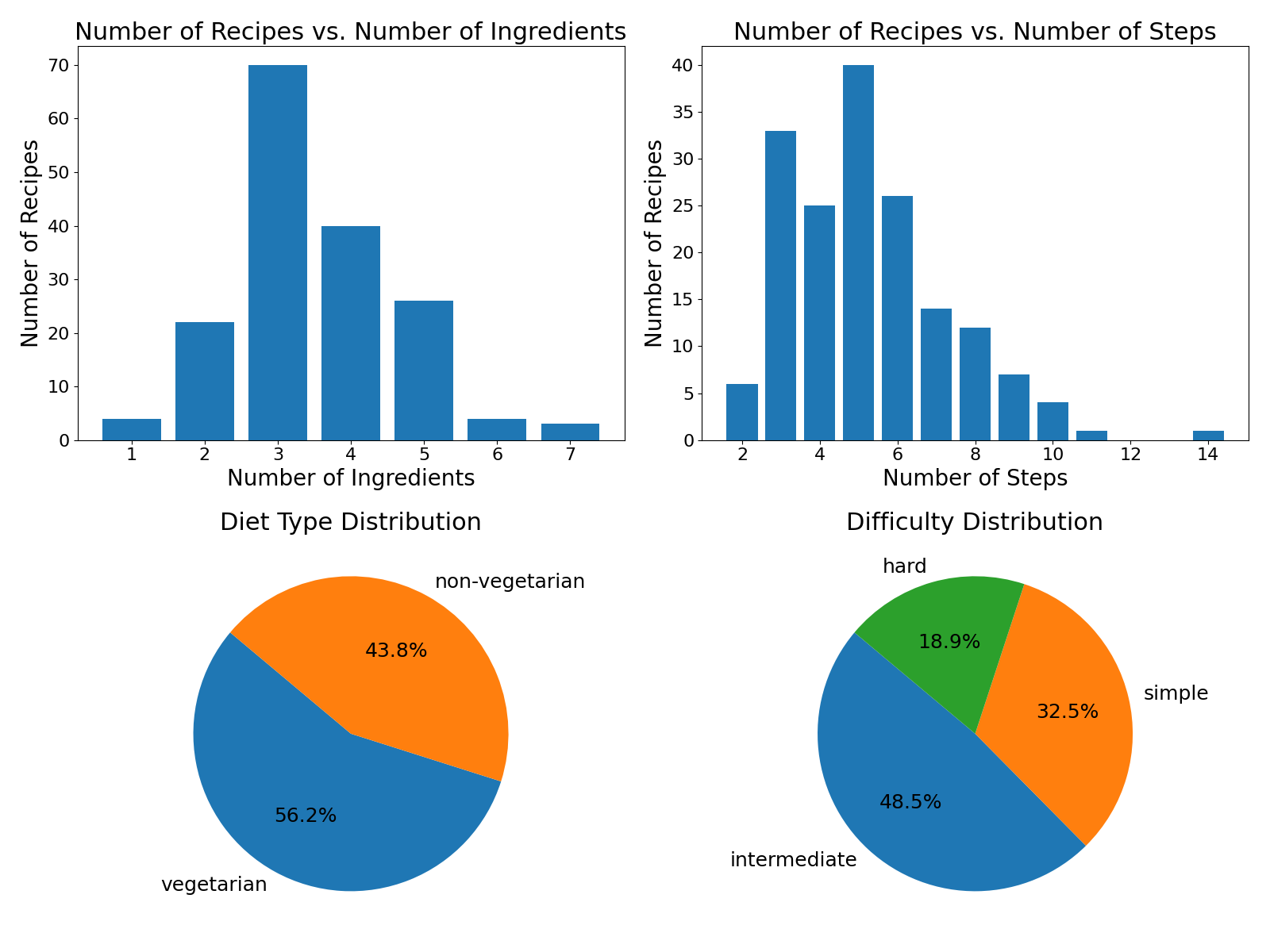}
    \caption{More data statistics of the indoor setting}
    \label{fig:indoor_statistics}
\end{figure}

In Figure \ref{fig:dataset_details}, we show the detailed distribution of all tasks. In Figure \ref{fig:indoor_statistics}, we show more statistics of the indoor cooking task, including the number of ingredients the task takes, the number of recipe steps as well as the distribution of diet types and difficulty levels.

\section{More Details about Data Collection}

\subsection{Outdoor Data Collection}

\textbf{Task intents and templates.} To better organize and summarize task intents, we’ve designed a set of task templates. Each template corresponds to several specific tasks. Below is an overview of these task templates.

\begin{tcolorbox}[breakable,title=Templates for Outdoor Tasks]
\textbf{Navigation:}\\
*  Show me the fastest route from \{origin\} to \{destination\}.\\
*  Plot a walking path from \{origin\} to \{destination\}, avoiding highways.\\
*  Starting at \{origin\}, guide me step-by-step to \{destination\}.\\
*  If I leave \{origin\} right now, what's the quickest way to reach \{destination\}?\\
*  ...\\
\\
\textbf{Shopping:}\\
*  Add all the items with \{\{quality\}\} on this page into my cart.\\
*  Add something like the \{\{item\}\} to my shopping cart.\\
*  Buy me a \{\{product\}\} with \{\{detail\}\}.\\
*  Can you add \{\{item\}\} \{\{condition\}\} to my wishlist?\\
*  How many calories are in \{\{item\}\} and \{\{secondary\_item\}\}? I need to select a lower one.\\
*  What size of \{\{item\}\} should I buy if \{\{condition\}\}?\\
*  ...\\
\\
\textbf{Wikipedia:}\\
*  Search for "\{query\}" and tell me more about it.\\
*  I would like to know more about "\{query\}".\\
*  Look up "\{query\}".\\
*  Provide me more information about "\{query\}".\\
*  ...
\end{tcolorbox}

In these templates, the placeholders in \{\} will be replaced with actual content. For example,
"Show me the fastest route from \{origin\} to \{destination\}" might become "Show me the fastest route from the Penn Station to Times Square".

\textbf{Task and location generation.} Although we have a list of templates for task intents, we still need to prompt the VLM to generate the initial task intent. Along with the task intent, since our dataset is a combination of embodied and web task, we also need to generate the location from our outdoor environment then we can proceed to the next step. The prompt to generate task and location can be seen in Section \ref{sec:outdoor_prompt}.

\begin{figure}[htbp]
    \centering
    \includegraphics[width=1\linewidth]{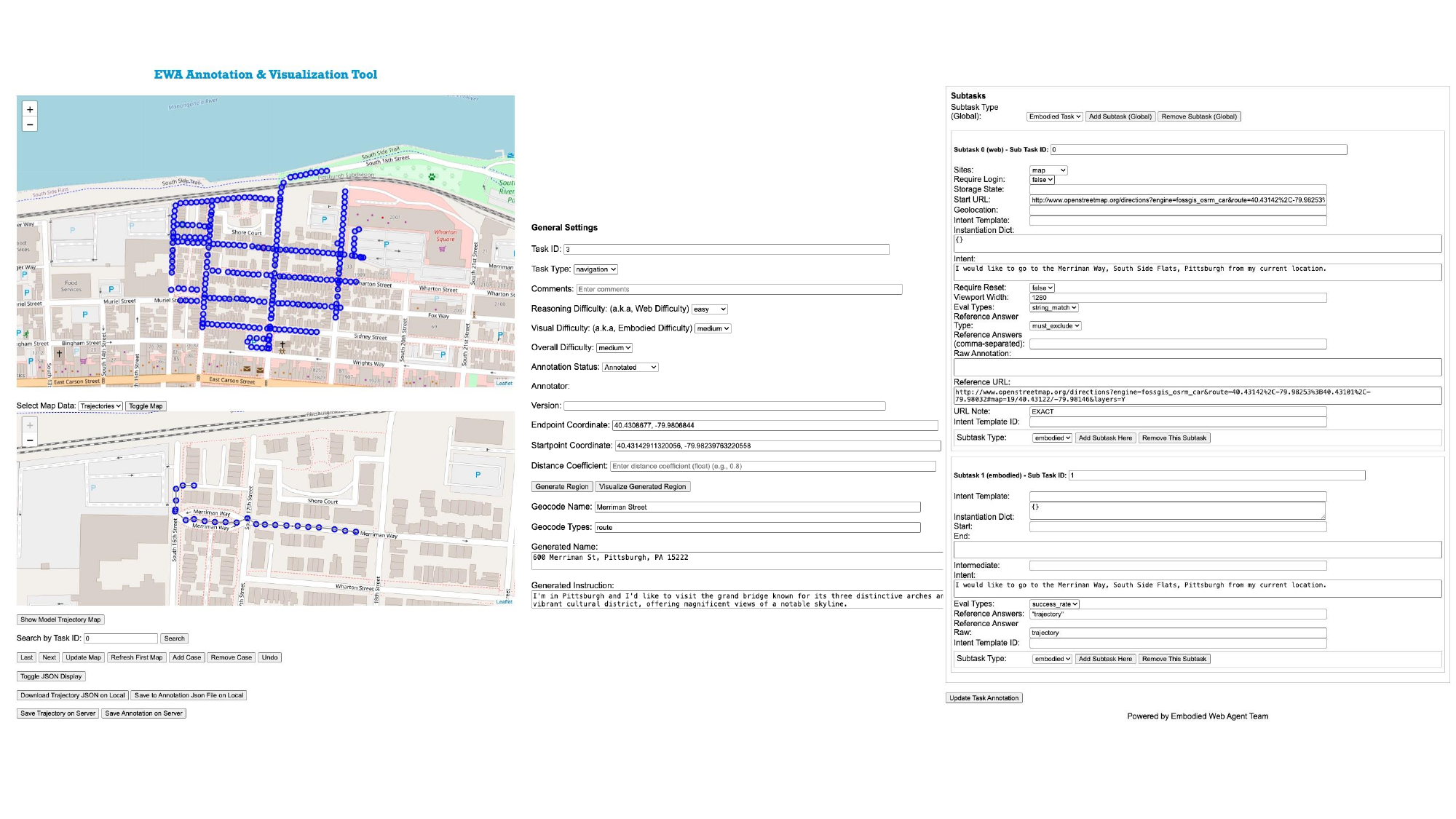}
    \caption{Annotation tool. Here is the annotation tool interface. It features three map windows side by side: one visualizes the coordinate points within the outdoor environment, the second shows the ground-truth outdoor navigation path, and the third displays the agent’s actual trajectory after navigation. Beyond simply visualizing points, you can directly edit the annotation JSON right in this interface. We load the JSON’s contents into the front end; once human verification is complete, any needed edits can be applied by clicking "Update Task Annotation", which pushes your changes back into the backend JSON file. This gives you both a clear visual overview and a streamlined labeling-and-verification workflow.}
    \label{fig:annotation_tool}
\end{figure}

\textbf{Annotation Tool.} To better support data annotation and visualization, we designed and built our own annotation tool shown as Figure \ref{fig:annotation_tool}. Having this graphical interface makes manual inspection and correction much simpler. Moreover, because our tasks involve outdoor navigation, we frequently need to visualize trajectories on a map, which lets us view the results in a very intuitive way. We can also directly update the data by using the annotation tool. When we modify the data, we save the changes made in the annotation tool’s interface directly to the backend JSON file.

\subsection{Geolocation Data Collection}
\textbf{Dataset Curation.} 
The dataset we present is composed of samples from the Breadth dataset of FairLocator. We randomly sample 142 locations and manually review each to ensure they are reasonable—meaning they contain enough visual cues for a model to make a prediction.

\textbf{Image Observations.}
To collect image observations, we use the Google Street View API to obtain views at our ground-truth coordinates. To support exploration in our embodied pipeline, we also query nearby "adjacent" viewpoints—defined as nodes within one edge in the Street View panorama graph, typically corresponding to a small translation from the initial location. For this reason, during sampling, we also exclude locations without adjacent viewpoints.

\textbf{Website Queries.} 
We utilize VisualWebArena to do the website queries. Since the VisualWebArena environment requires configuration files specific to each query or intent, we dynamically generate a new configuration file each time the agent creates a new search prompt. We do not use predetermined configuration files, as we want to evaluate the agent's ability to use visual cues and identify its own knowledge gaps. Thus, the queries and configurations for each run are random and unique to the model in some sense. We also stipulate that queries should be styled simplistically and be optimized for Wikipedia searches since we only allow the model to access the Wikipedia site within our web environment. We concatenate these queries and their results in a context cache to feed to the agent during confidence estimations and the final prediction.

\section{Qualitative Examples}
\subsection{Outdoor Planning}
The outdoor planning consists of three core subtask types, that are navigation, traveling, and shopping. Here, we present four illustrative examples for navigation error, traveling success, shopping error, and shopping success.

\begin{figure}[htbp]
  \centering
  \includegraphics[width=1.0\textwidth]{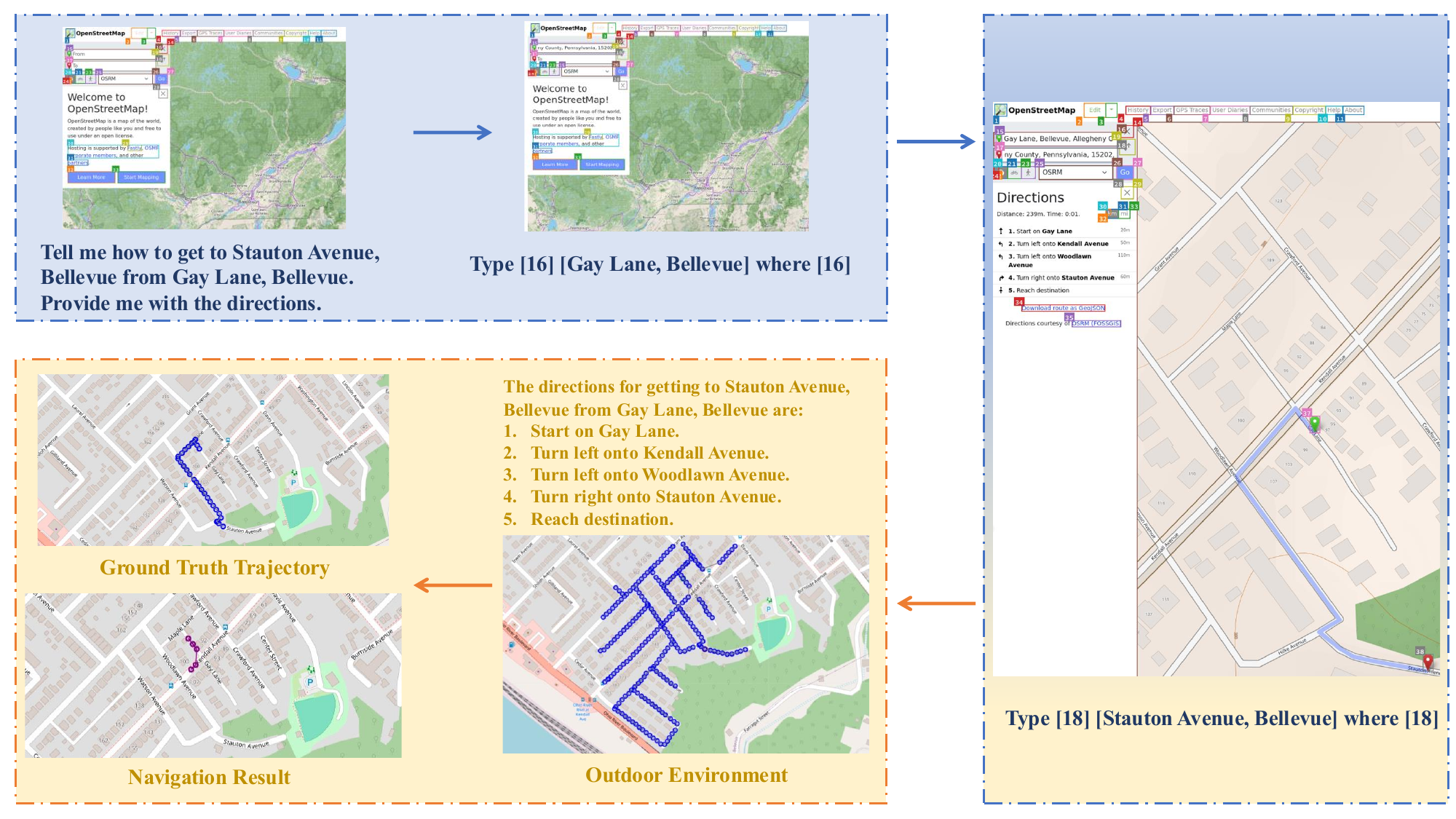}
  \caption{Navigation error. The agent’s failure to correctly understand the directions from the map website led to navigation errors in the outdoor environment.}
  \label{fig:navigation_error}
\end{figure}

In Figure \ref{fig:navigation_error}, the agent misinterprets complex map directions and moves in the wrong direction within the outdoor environment, ultimately causing the navigation task to fail. This highlights the agent’s current limitations and biases when reading, processing, and grounding intricate routing information in the real world.

\begin{figure}[htbp]
  \centering
  \includegraphics[width=1.0\textwidth]{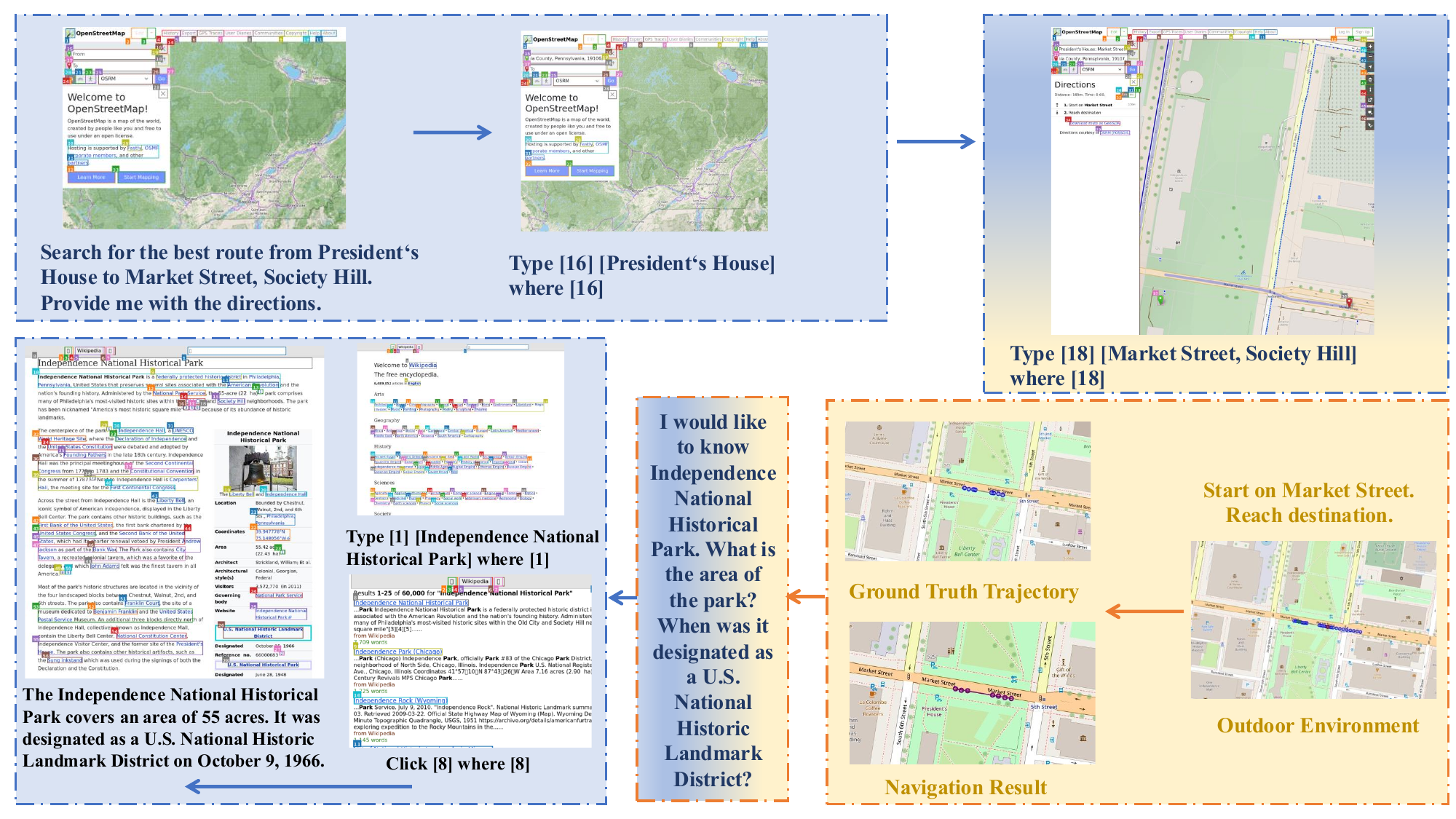}
  \caption{Traveling success. The agent first correctly understood the user’s request and provided accurate map directions. It then navigated through the outdoor environment and moved to the correct location. Because this was a traveling task at a tourist site, the agent finally queried the environment by consulting the Wikipedia page, obtained the right information, and successfully completed the entire task.}
  \label{fig:traveling_success}
\end{figure}

In Figure \ref{fig:traveling_success}, it is a representative traveling task that the agent correctly processes the directions, reaches the designated location. Then, since the location is a point of interest, the agent queries Wikipedia and retrieves the correct information. This case demonstrates how the agent seamlessly integrates information from both web sources and the embodied environment to complete the entire traveling workflow.

\begin{figure}[htbp]
  \centering
  \includegraphics[width=1.0\textwidth]{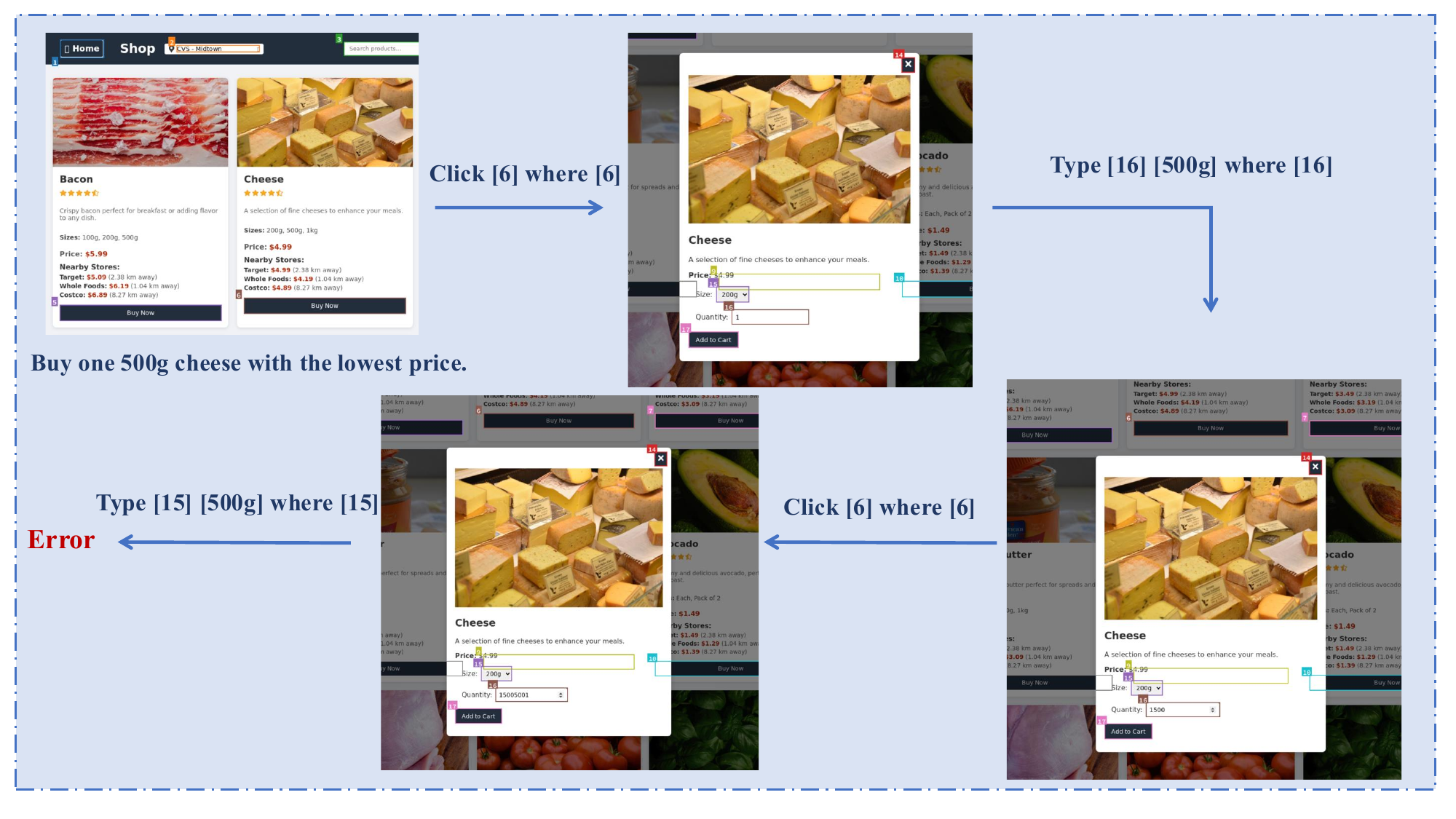}
  \caption{Shopping error. The agent failed to correctly interpret the elements on the webpage and likewise did not produce the correct action based on its visual and language inputs (it should have clicked instead of typing), which ultimately caused the agent to err and fail to complete the shopping task.}
  \label{fig:shopping_error}
\end{figure}

In Figure \ref{fig:shopping_error}, the agent fails to understand the function of certain web elements and, based on its vision-language input, does not ground its decision to the correct action (clicking versus typing). This failure exposes areas for improvement in the agent’s action grounding capabilities, especially for web interactions. It also reveals weaknesses in its visual grounding when the clickable target is not visually salient.

\begin{figure}[htbp]
  \centering
  \includegraphics[width=1.0\textwidth]{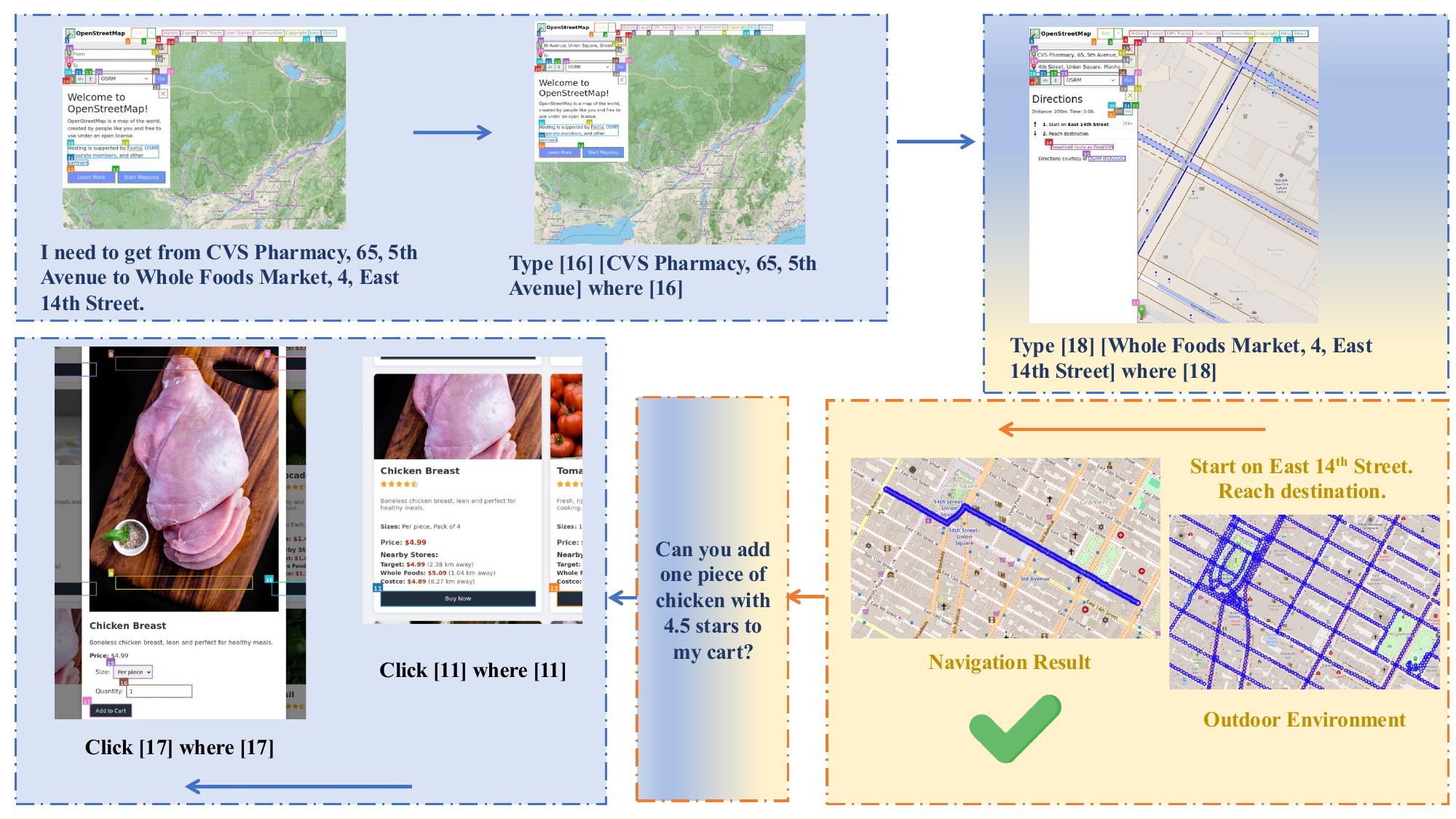}
  \caption{Shopping success. The agent successfully completed all of its subtasks. First, it retrieved the correct directions from the webpage, then navigated to the right location and began shopping. Finally, during the shopping process, it selected the correct item and saw the entire shopping task through to completion.}
  \label{fig:shopping_success}
\end{figure}

In Figure \ref{fig:shopping_success}, the agent successfully navigates to the correct store, processes the online shopping interface, and ultimately selects and purchases the right product. This success case illustrates the agent’s ability to coordinate web-derived information with real-world movement to complete the full shopping task.

Together, these examples vividly illustrate the challenges and progress in grounding web information within embodied tasks across navigation, traveling, and shopping scenarios.

\subsection{Indoor Cooking}
\begin{figure}
    \centering
    \includegraphics[width=1\linewidth]{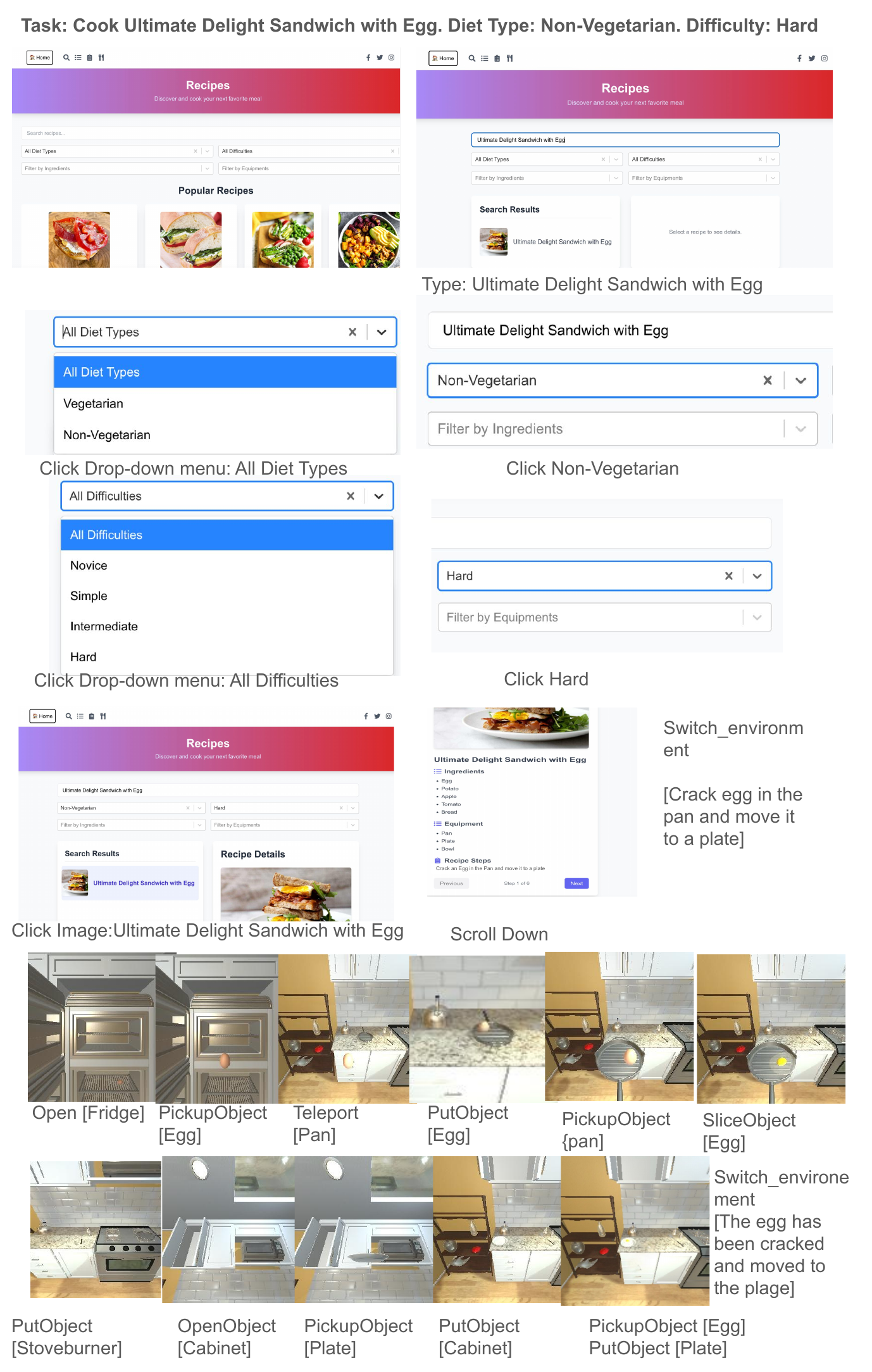}
    \caption{Qualitative example of indoor cooking - Part 1}
    \label{fig:indoor_1_1}
\end{figure}

\begin{figure}
    \centering
    \includegraphics[width=1\linewidth]{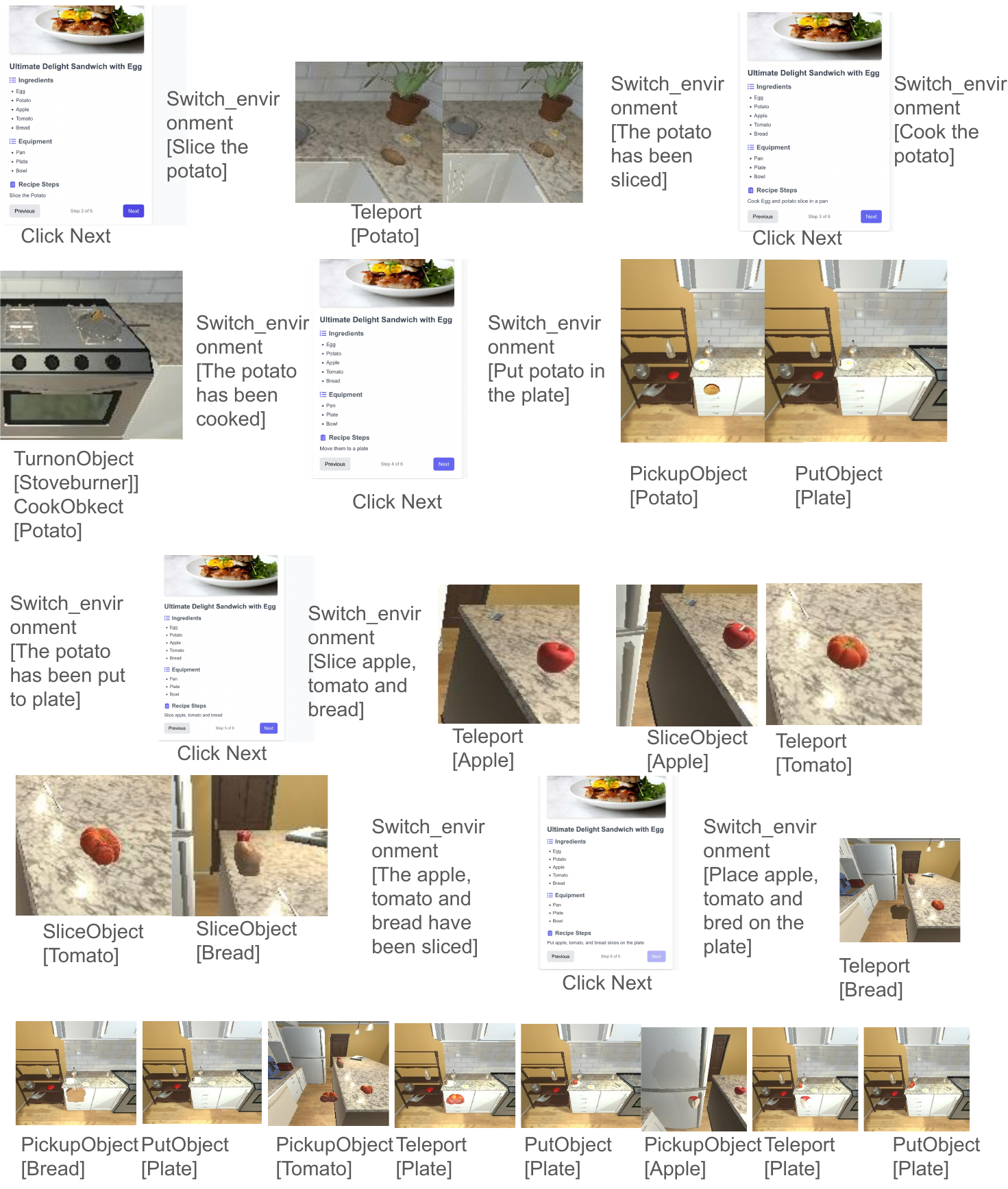}
    \caption{Qualitative example of indoor cooking - Part 2}
    \label{fig:indoor_1_2}
\end{figure}

\begin{figure}
    \centering
    \includegraphics[width=1\linewidth]{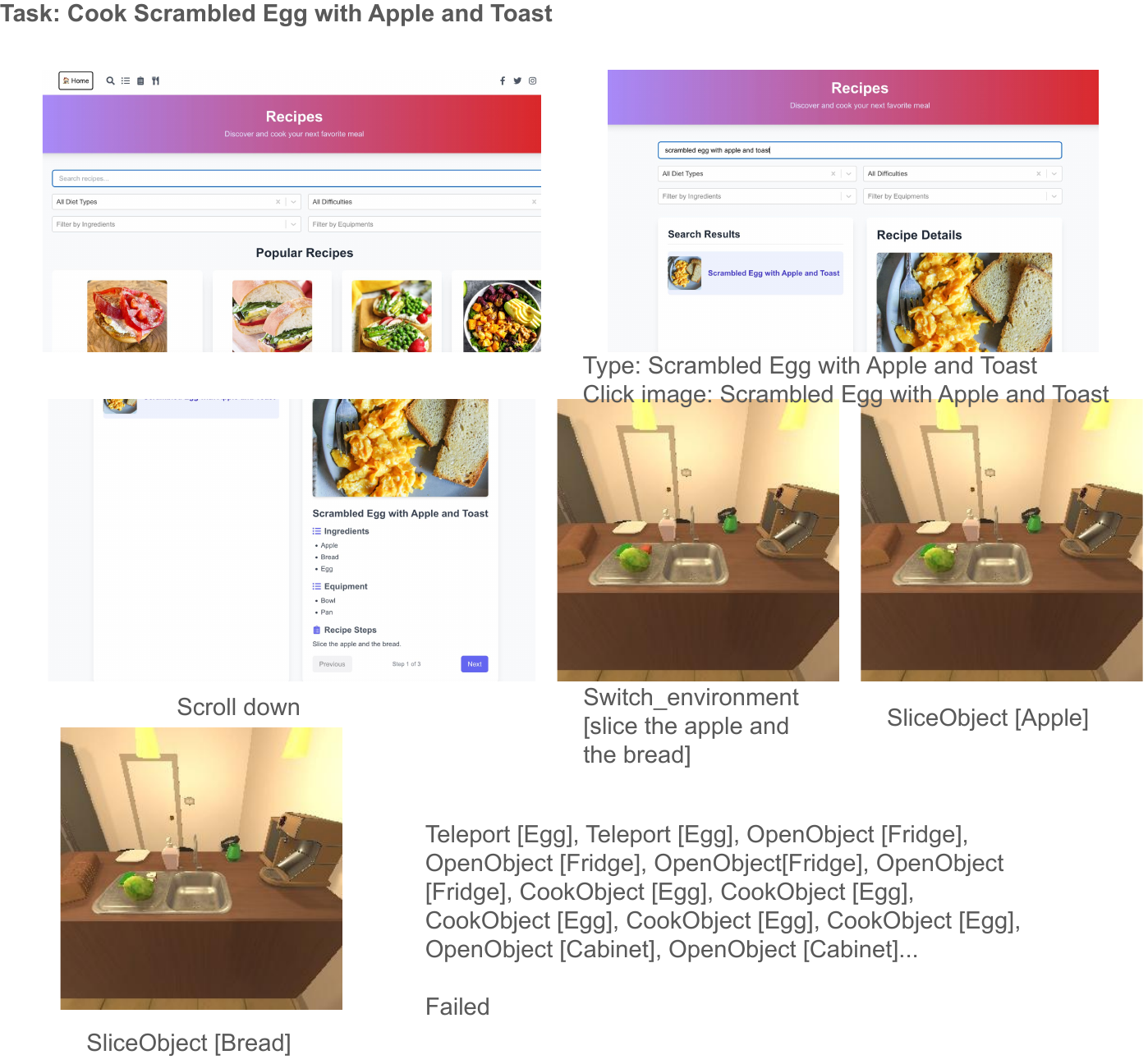}
    \caption{Failure case of indoor cooking}
    \label{fig:indoor_failure}
\end{figure}

We show a full example of carrying out a cooking task following web instructions in Figure \ref{fig:indoor_1_1} and \ref{fig:indoor_1_2}. As we can see, the model needs to perform multi-step iterative reasoning between the web side and embodied side to complete a complex cooking task. In Figure \ref{fig:indoor_failure}, we show a failure case. It fails because: 1) action grounding error. The web instruction is to slice apple and bread. However, it also tries to crack the egg. 2) Stuck in the embodied side and cannot go back to the web side. When it fails to crack an egg, it starts to perform random actions in the embodied environment without trying to go back to the web environment.

\subsection{Geolocation}
\begin{figure}[htbp]
    \centering
    \includegraphics[width=1\linewidth]{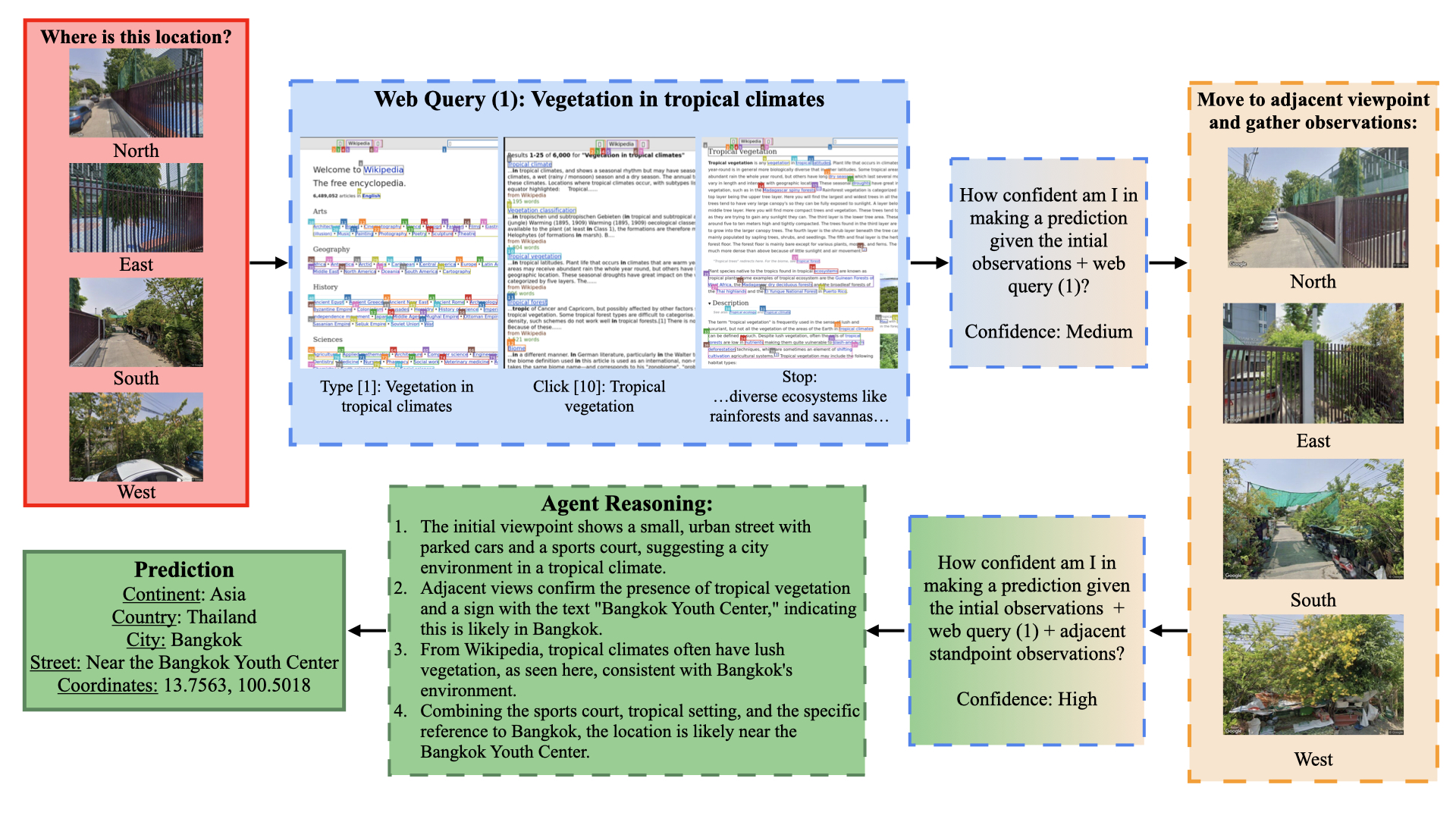}
    \caption{An exemplar pipeline of completing the geolocation task.
The red box indicates initial input, the blue box indicates web interaction, the orange box indicates embodied interaction, and the green boxes indicate agent reasoning and prediction. Boxes with gradient colors indicate switching from one environment to the other.}
    \label{fig:geolocation_example}
\end{figure}

Figure \ref{fig:geolocation_example} illustrates a step-by-step example of the embodied geolocation pipeline. We begin with images from the initial standpoint—one image facing each of the four cardinal directions. The agent uses these observations to generate a web query, formulated in the style of a Wikipedia search. 
This query is executed using the VisualWebArena environment and the resulting web page content is retrieved. Both the image observations and the web search results are then passed to the agent's confidence estimation module, which assesses whether the current context is sufficient for making an accurate geolocation prediction. 
In this example, the agent initially determines that it lacks sufficient confidence. It then chooses to move to another nearby adjacent standpoint, gathers new image observations, and re-evaluates its confidence. Upon receiving the additional context, the agent is confident enough to make a prediction and proceeds to output both its predicted location and its reasoning.

\section{LLM Prompts}
\subsection{Outdoor}\label{sec:outdoor_prompt}
Here we list three key prompts used in our outdoor tasks. First, the prompt for generating locations at the very beginning. Second, based on those generated locations and the initial instruction, we use a task generation prompt to create the detailed subtasks, both embodied and web-based. After generating each task, we perform human verification and the verified data then becomes the dataset we use for our experiments. Once the experiments begin, the web‐interaction portion still relies on the same prompts used in VisualWebArena. For the outdoor navigation portion, we employ the outdoor navigation prompt shown below. For visualization purposes, some of the prompts shown have been appropriately shortened.

\begin{tcolorbox}[breakable,title=Location Generation for Outdoor Tasks]
You are an AI assistant that is familiar with cities all around the world. For a given city, please provide an iconic locations. For each location, provide a navigation instruction that captures its unique characteristics, historical significance, cultural importance, or architectural features WITHOUT directly mentioning its name. The navigation instructions should be specific enough that someone knowledgeable about the city could identify the exact location. Return the results as a JSON list where each element contains 'location' (a searchable address for geocoding) and 'instruction' (an informative instruction that avoids using the location's name but will uniquely locate the location). REMEMBER that the location MUST be in the city of New York, USA; Philadelphia, USA; Boston, USA; Pittsburgh, USA! \\
\\
Here are four examples: 
\\ 
\texttt{[{"location": "350 5th Ave, New York, NY 10118", "instruction": "I'm in New York City and I'd like to go to the Art Deco skyscraper from the 1930s that held the title of world's tallest building for nearly 40 years. It has 102 stories and is an enduring symbol of the city's ambition."}]}
\\
\texttt{[{"location": "520 Chestnut St, Philadelphia, PA 19106", "instruction": "I'm in Philadelphia and I'd like to go to the red-brick Georgian hall with a white steeple in the historic district where revolutionary delegates gathered in the 18th century to debate and adopt the nation's founding documents."}]}
\\
\texttt{[{"location": "4 Jersey St, Boston, MA 02215", "instruction": "I'm in Boston and I'd like to go to the century-old ballpark that opened in 1912, famous for its emerald-green left-field wall and as the longstanding home of one of Major League Baseball's oldest franchises."}]}
\\
\texttt{[{"location": "601 Commonwealth Pl, Pittsburgh, PA 15222", "instruction": "I'm in Pittsburgh and I'd like to go to the 150-foot-tall water jet fountain at the tip of downtown's triangular park, marking where three rivers converge against a backdrop of the city skyline."}]"}
\\
\\
Only give me one spot in one of these cities: New York, USA; Philadelphia, USA; Boston, USA; Pittsburgh, USA.
\end{tcolorbox}

\begin{tcolorbox}[breakable,title=Task Generation for Outdoor Tasks]
You are an Embodied Web Agent capable of obtaining information from webpages and executing tasks within an embodied environment. I will provide you with a generated\_instruction and a generated\_name from the embodied environment. The generated\_instruction is a description of the task that outlines what the embodied agent needs to do, while the generated\_name is the name or address of a location that the embodied agent needs to go to.\\
\\
Based on the generated\_instruction and generated\_name, you need to generate tasks that both the web and the embodied agent must execute. These tasks should ensure that after execution, the embodied web agent can obtain information from the web to assist in completing the task in the embodied environment.\\
\\
We have four types of webpages, and I will provide you with a task category. Based on the task category, you need to generate one or more tasks that interact with the webpages. The tasks must make use of at least one of the following webpages (or multiple): Task categories include: Shopping, Navigation, and Traveling.\\
\\
The descriptions for these task categories are as follows: \\
1.	Shopping: You need to search for product information, prices, store locations on the web. Then compare this information to find the most suitable store. Finally, the outdoor embodied agent can use the store address from the web to reach the store.\\
2.	Navigation: You need to search for maps and route planning on the web. These details will help the outdoor embodied agent find the best route from the current location to the destination.\\
3.	Traveling: You need to search for tourist attractions, travel guides, local culture, etc., on the web. Then, this information will help the outdoor embodied agent plan an itinerary and choose attractions or activities.\\
\\
The types of webpages include: Shopping, OpenStreetMap, Wikipedia, and Homepage.\\
Here are descriptions of these webpages:\\
1.	Shopping: This is a shopping website that provides information on various products, including prices and store locations. You can look for detailed product information and purchasing options here.\\
2.	OpenStreetMap: This is an OpenStreetMap website, which provides maps and route planning services. You can search for your current location, destination, and best routes here.\\
3.	Wikipedia: This is a Wikipedia website that provides encyclopedic knowledge on various topics. You can look up tourist attractions, local culture, travel guides, and more.\\
4.	Homepage: This is a homepage website that provides links to the above websites. These websites can also lead you back to this homepage, making it convenient for users to switch between different websites.\\
\\
Below are three examples. You need to generate output in this format:
\\
Example 1:\\
Task Category: Traveling\\
generated\_instruction: I'd like to visit the iconic Central Park, a sprawling urban park in New York City, known for its picturesque landscapes, recreational activities, and cultural landmarks.
generated\_name: Central Park, New York, NY\\
web\_task\_intent\_0: Search for information about Central Park on the Wikipedia website. I would like to know about its open hours.\\
embodied\_task\_intent\_1: I would like to explore Central Park and its various attractions.\\
web\_task\_intent\_2: Find the best walking route between my current location and Central Park using the OpenStreetMap website.\\
\\
Example 2:\\
Task Category: Navigation\\
...\\
\\
Example 3:\\
Task Category: Shopping\\
...\\
\\
You are given the following inputs: \\
Task Category: [task\_category\_placeholder]\\
generated\_instruction: [generated\_instruction\_placeholder]\\
generated\_name: [generated\_name\_placeholder]\\
\\
Please only generate the embodied\_task\_intent and web\_task\_intent below.
\end{tcolorbox}

\begin{tcolorbox}[breakable,title=Outdoor Navigation]
You are an embodied navigation agent operating within a street-view graph environment.\\
\\
Each environment is defined by:\\
(1) A source node (starting latitude-longitude).\\
(2) A target node (destination latitude-longitude).\\
(3) A set of graph nodes, each with:
\begin{enumerate}[%
  label=(\alph*), 
  leftmargin=2em,       
  labelsep=0.5em,       
  itemsep=0pt,          
]
\item A unique node ID (lat-lng string).
\item Four street-view images (north, east, south, west) as your visual observations.
\item A list of neighbor nodes with absolute heading, descriptive text, and edge distance.
\end{enumerate}
Your Objective:\\
Navigate step-by-step from the source to the target node by selecting exactly one neighbor at each step, according to the given parsed navigation instructions and the visual/textual context.\\
\\
Available Inputs:\\
(1) Current node ID (string).\\
(2) Target node ID (string).\\
(3) Current absolute heading (degrees clockwise from true north).\\
(4) Parsed instructions: a list of {action, distance} pairs (action $\in$ {straight, left, right}).\\
(5) Remaining distance (meters) to complete the current instruction step.\\
(6) List of previously visited node IDs (to avoid loops).\\
(7) For each neighbor:
\begin{enumerate}[%
  label=(\alph*), 
  leftmargin=2em,       
  labelsep=0.5em,       
  itemsep=0pt,          
]
\item Neighbor ID.
\item Absolute heading (°).
\item Relative heading to your current facing (°).
\item Distance (m) along the edge.
\end{enumerate}
(8)Four visual observations: street-view images facing north, east, south, and west.\\
\\
Your Task:\\
Based on all of the above, choose exactly one neighbor ID that best:\\
1. Follows the current action instruction (straight/left/right) relative to your facing.\\
2. Moves you toward the target by reducing distance.\\
3. Does not revisit an already visited node.\\
\\
Response Format:\\
Reply with exactly one node ID (lat-lng string) on a single line, with no additional commentary.
\end{tcolorbox}

\subsection{Geolocation}
We design separate prompting strategies for the baseline and embodied pipelines. Additionally, we found that Qwen required significantly stricter prompt constraints to produce output consistent with our expected format, so we created dedicated prompts for Qwen.

\begin{figure}[htbp]
  \centering

  \begin{subfigure}[b]{0.48\textwidth}
    \includegraphics[width=\linewidth]{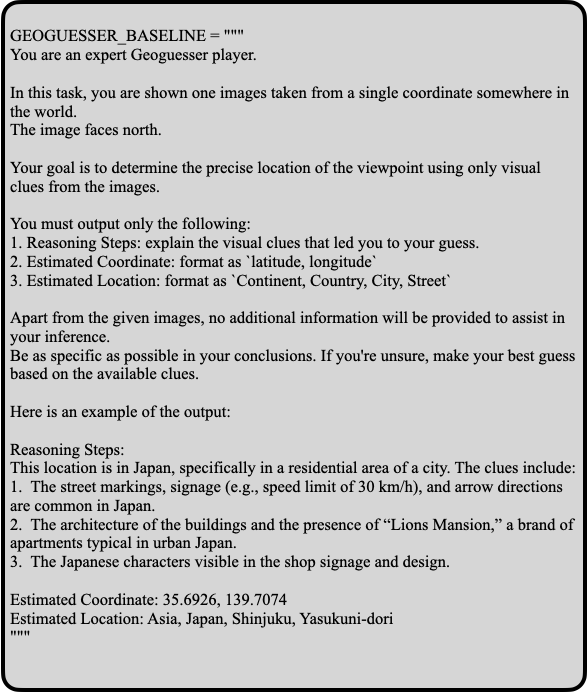}
    \caption{GPT, Gemini baseline prompt.}
    \label{fig:subfig1}
  \end{subfigure}
  \hfill
  \begin{subfigure}[b]{0.48\textwidth}
    \includegraphics[width=\linewidth]{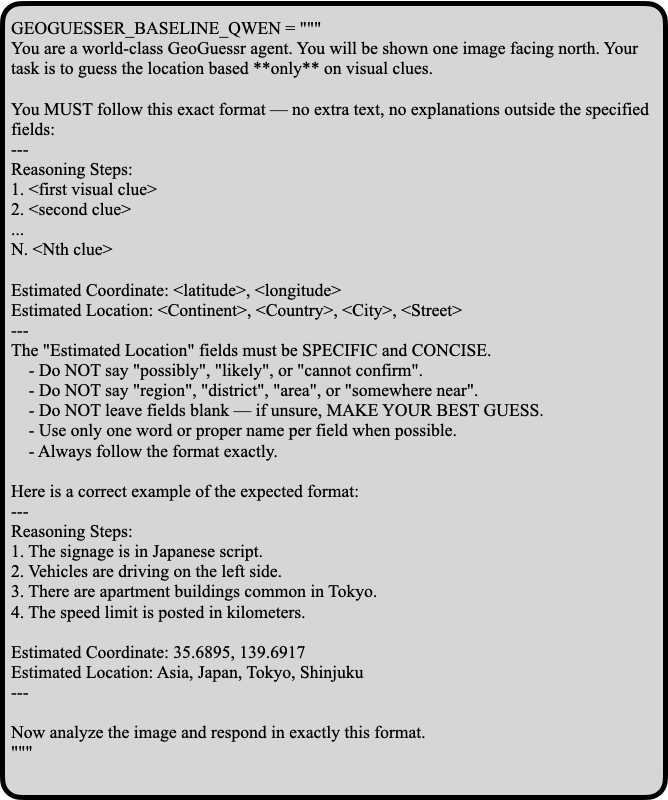}
    \caption{Qwen baseline prompt.}
    \label{fig:subfig2}
  \end{subfigure}

  \caption{LLM prompts for geolocation baseline.}
  \label{fig:geolocation_baseline}
\end{figure}

In the baseline setting, we only prompt the vision-language model (VLM) with a single north-facing image from the initial standpoint. As shown in Figure \ref{fig:geolocation_baseline}, we use one prompt for GPT and Gemini and a separate version tailored to Qwen.

\begin{figure}[htbp]
  \centering

  \begin{subfigure}[b]{0.48\textwidth}
    \includegraphics[width=\linewidth]{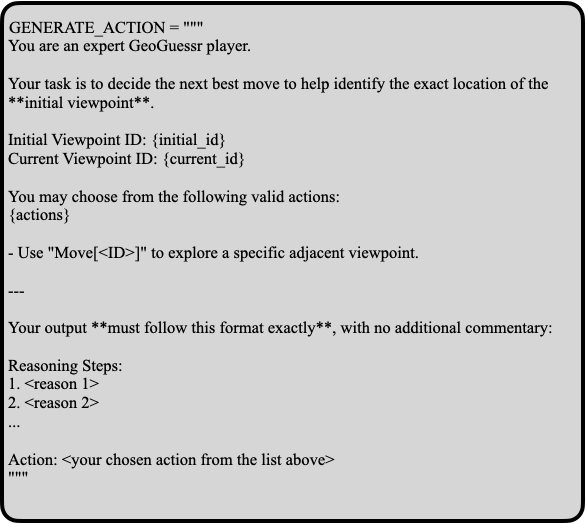}
    \caption{Action generation prompt.}
    \label{fig:subfig1}
  \end{subfigure}
  \hfill
  \begin{subfigure}[b]{0.5\textwidth}
    \includegraphics[width=\linewidth]{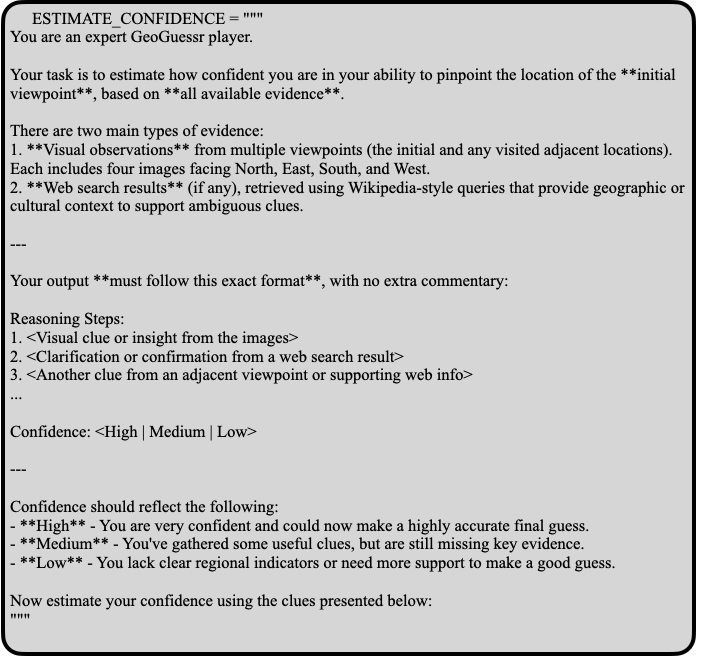}
    \caption{Confidence estimation prompt.}
    \label{fig:subfig2}
  \end{subfigure}

  \begin{subfigure}[b]{0.48\textwidth}
    \includegraphics[width=\linewidth]{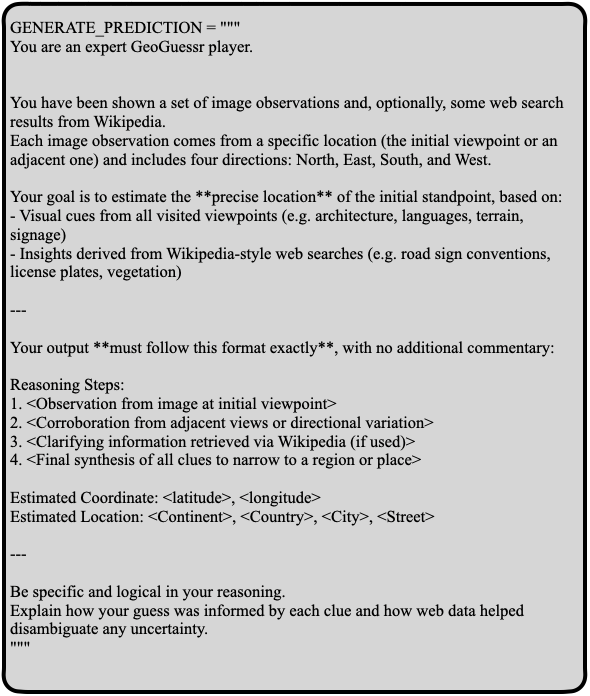}
    \caption{GPT, Gemini prediction prompt.}
    \label{fig:subfig3}
  \end{subfigure}
  \hfill
  \begin{subfigure}[b]{0.5\textwidth}
    \includegraphics[width=\linewidth]{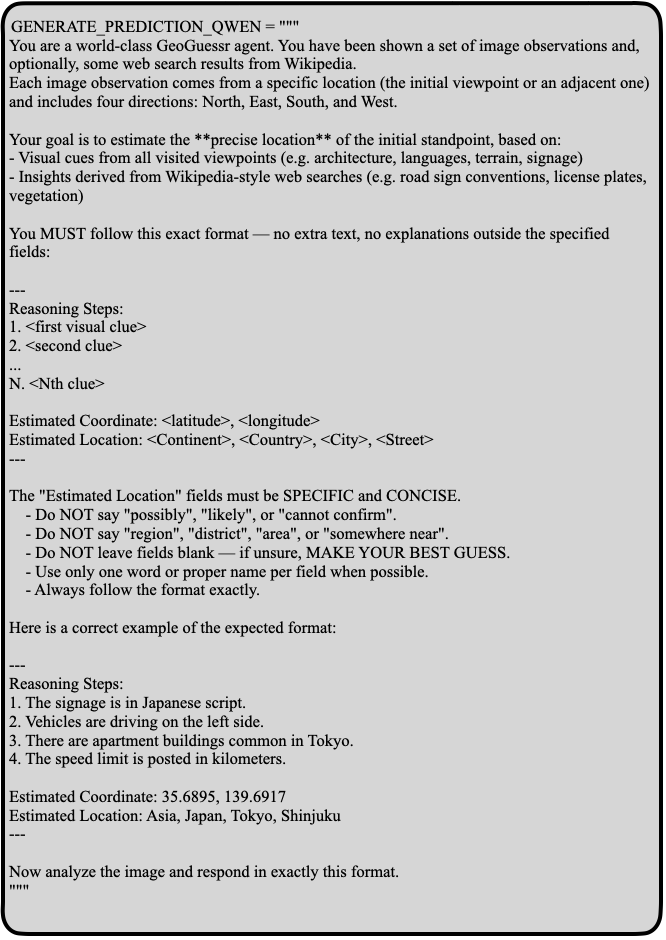}
    \caption{Qwen prediction prompt.}
    \label{fig:subfig4}
  \end{subfigure}

  \caption{LLM prompts for geolocation task exploration, confidence estimation, and final predictions.}
  \label{fig:embodied_geolocation}
\end{figure}

\begin{figure}[htbp]
    \centering
    \includegraphics[width=0.7\linewidth]{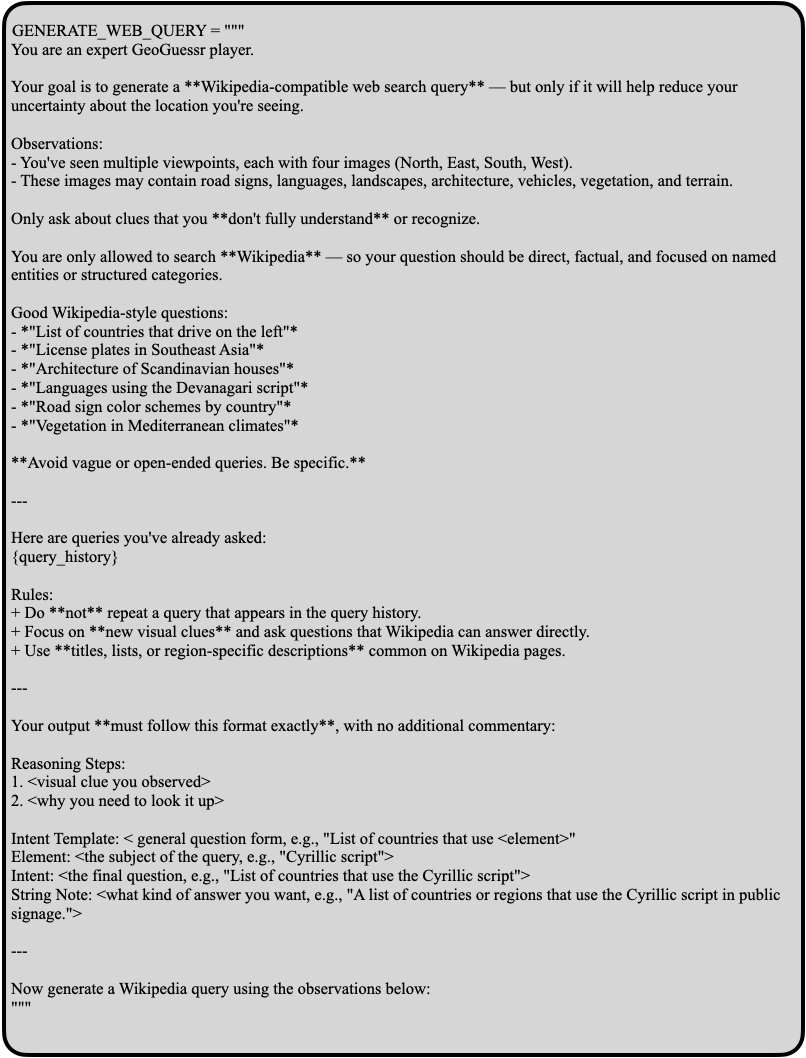}
    \caption{LLM prompt for geolocation task web query generation.}
    \label{fig:geolocation_web}
\end{figure}

In contrast, the embodied pipeline involves multiple types of prompts, as illustrated in Figure \ref{fig:embodied_geolocation}. We use distinct prompts to:
\begin{itemize}
    \item instruct the agent to move to adjacent standpoints,
    \item estimate confidence based on the current context,
    \item and generate a final location prediction.
\end{itemize}
The web query prompt (Figure \ref{fig:geolocation_web}) is issued after each new round of observations. The web query is executed in VisualWebArena and we add the results to a growing context cache along with prior image observations and web results. This evolving context is provided to the agent for both confidence estimation and the final prediction.

\section{Web Environment}
\begin{figure}
    \centering
    \includegraphics[width=1\linewidth]{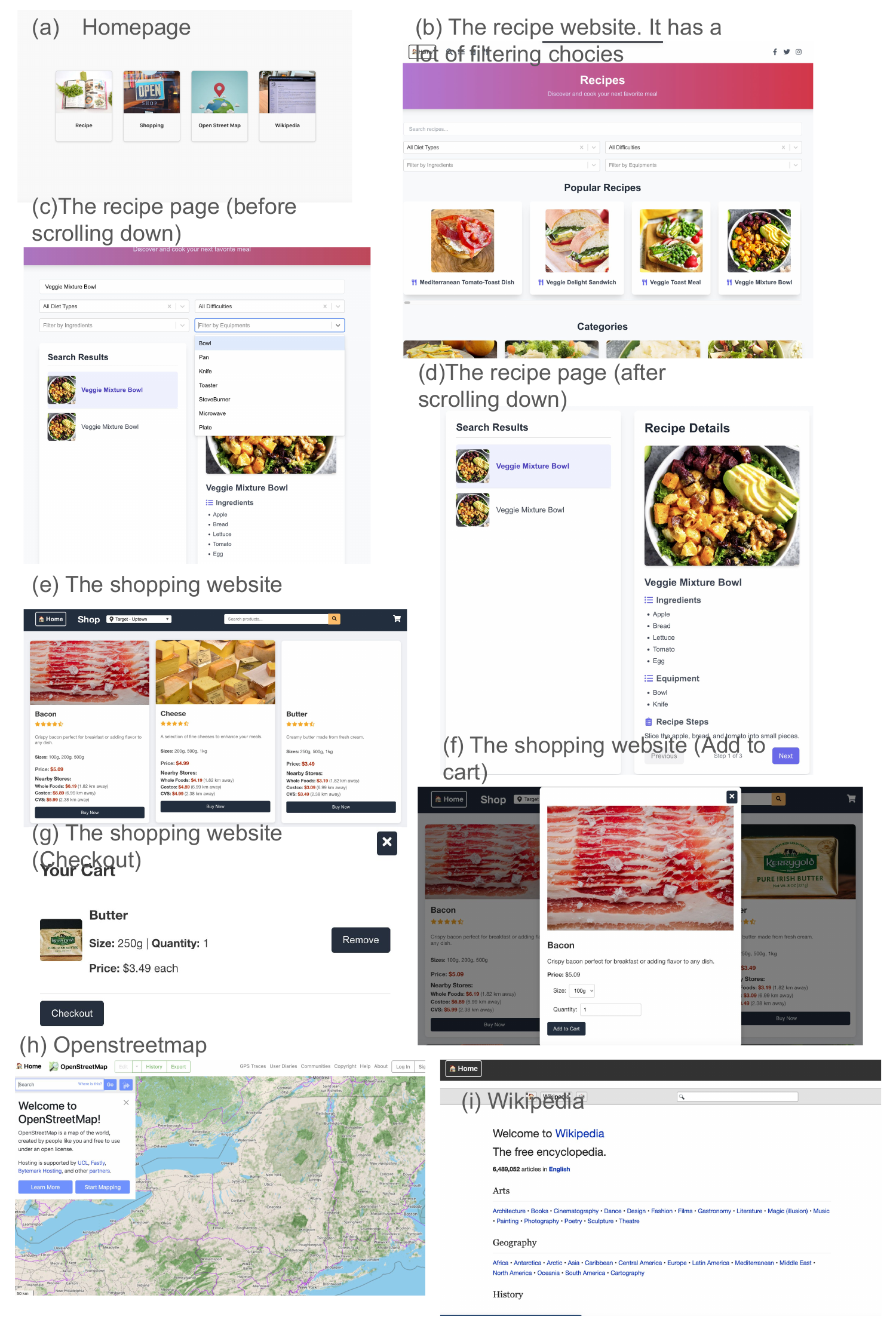}
    \caption{Web environment screenshots}
    \label{fig:web_env}
\end{figure}

In Figure \ref{fig:web_env}, we show screenshots of our web environment. Please go to \url{http://98.80.38.242:1220/} for more details.

\section{Human Performance}
To establish meaningful benchmarks for our evaluations, we recruited undergraduate and graduate student volunteers from UCLA's Computer Science and Statistics departments. Participants were selected to represent a diverse range of technological familiarity and task-specific expertise. Each volunteer participated in a 2-hour session where they completed the same set of tasks that were presented to the AI models, covering both web-based and embodied scenarios in shopping, traveling, and cooking domains.

Our analysis reveals remarkable human performance across all domains, with overall accuracy rates ranging from 77.08\% to 92.59\%, significantly outperforming even the most capable AI systems. Particularly noteworthy is the consistent human performance across both web-based and embodied tasks, whereas AI models showed dramatic performance drops in embodied scenarios. This performance gap underscores the substantial challenges remaining in developing AI systems that can match human-level understanding and execution of everyday tasks that require multimodal reasoning, real-world knowledge application, and adaptive problem-solving strategies in response to environmental feedback.

\end{document}